\documentclass{zlab}

\usepackage[T1]{fontenc}
\usepackage{tgpagella}
\usepackage{mathpazo}
\usepackage{inconsolata}

\usepackage{xspace}
\usepackage{graphicx,amsmath,amssymb,hyperref}
\usepackage{adjustbox}
\usepackage[authoryear,round]{natbib}
\setcitestyle{aysep={,},yysep={;}}
\usepackage{xcolor}
\usepackage{booktabs}
\usepackage{nicefrac}
\usepackage{float}
\usepackage{subcaption}
\usepackage{makecell}
\usepackage{wrapfig}
\usepackage[section]{placeins}
\usepackage{needspace}
\usepackage{enumitem}
\usepackage{listings}
\usepackage{array}

\graphicspath{{figures/}}

\setlist[itemize]{leftmargin=*, itemsep=0.35em, topsep=0.35em}

\newcommand{\tablefontsize}{\fontsize{8pt}{9.5pt}\selectfont}


\definecolor{link}{HTML}{0063BE}
\hypersetup{
  colorlinks=true,
  urlcolor=magenta,
  linkcolor=black,
  citecolor=link,
  filecolor=black,
  pdfborder={0 0 0}
}

\providecommand{\equationname}{Equation}
\providecommand{\sectionname}{Section}
\newcommand{\figref}[1]{%
  \figurename~\hyperref[#1]{\textcolor{link}{\ref*{#1}}}%
}
\newcommand{\tabref}[1]{%
  \tablename~\hyperref[#1]{\textcolor{link}{\ref*{#1}}}%
}
\newcommand{\eqrefc}[1]{%
  \equationname~\hyperref[#1]{\textcolor{link}{\ref*{#1}}}%
}
\newcommand{\secrefc}[1]{%
  \sectionname~\hyperref[#1]{\textcolor{link}{\ref*{#1}}}%
}

\newcommand{\github}{\raisebox{-1.3pt}{\includegraphics[height=1.05em]{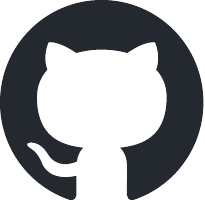}}\xspace}
\newcommand{\worldwideweb}{\raisebox{-1.3pt}{\includegraphics[height=1.05em]{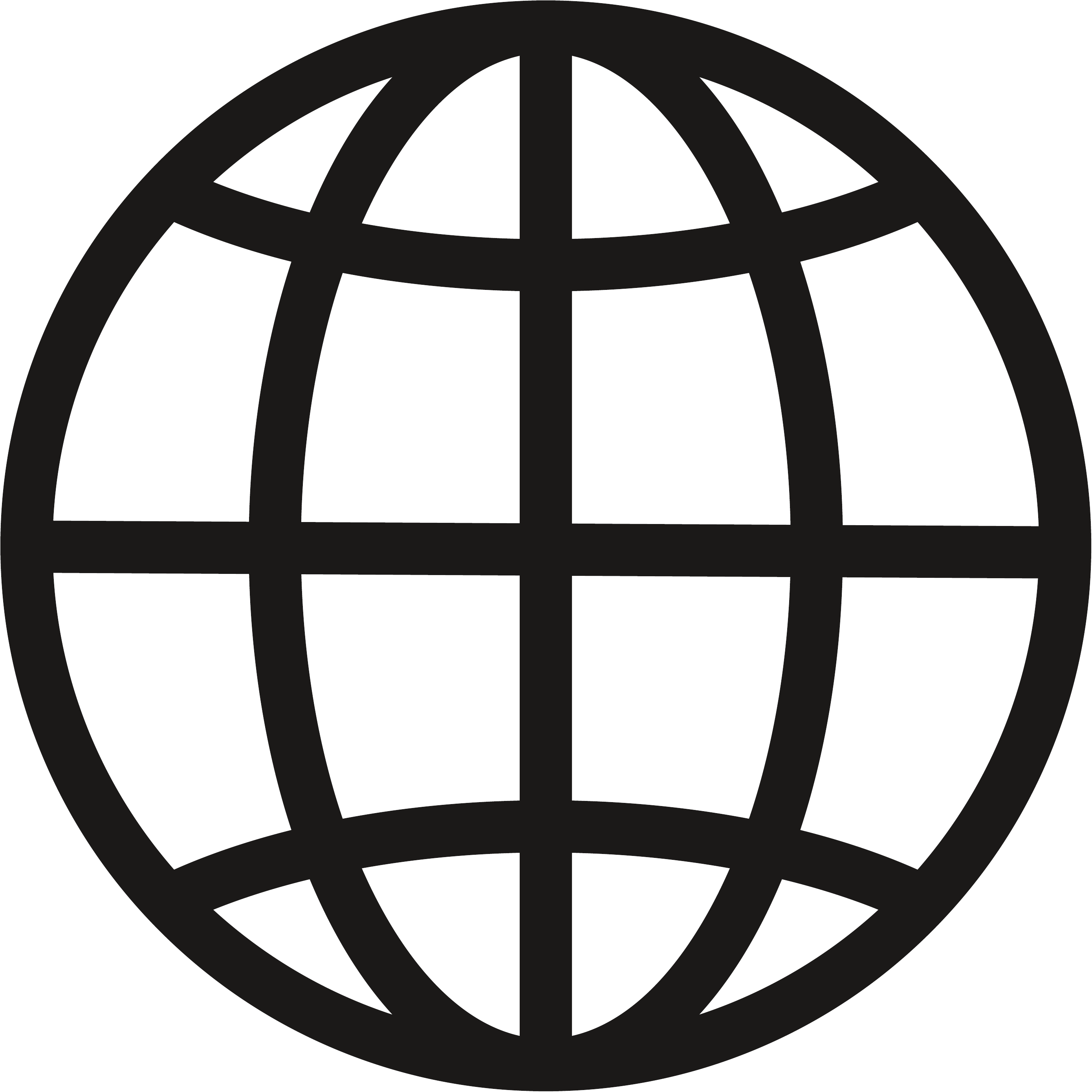}}\xspace}
\newcommand{\trajectoryicon}{\raisebox{0.08em}{\(\blacktriangleright\)}\xspace}

\makeatletter
\fancypagestyle{firststyle}{
  \fancyhead[L]{}
  \fancyhead[C]{}
  \fancyhead[R]{}
  \fancyfoot[L]{\footerfont \the\correspondingauthor}
  \fancyfoot[R]{}
}
\makeatother

\DeclareTextFontCommand{\textbf}{\bfseries}
\title{\centering \fontsize{18}{24}\selectfont{\textsc{CEO-Bench}: Can Agents Play the Long Game?}}

\author{%
    \vspace{.2cm}
    \parbox{\textwidth}{\centering
      Haozhe Chen \quad Karthik Narasimhan \quad Zhuang Liu\\
      Princeton University
    }
}

\makeatletter
\renewcommand{\abscontent}{}%
\makeatother

\newenvironment{abstractblock}{%
  {\centering\large\bfseries Abstract\par}
  \vspace{0.2em}
  \begin{list}{}{%
      \setlength{\leftmargin}{2em}
      \setlength{\rightmargin}{2em}
      \setlength{\topsep}{0pt}
      \setlength{\parsep}{0pt}
  }
  \item[]
}{%
  \end{list}
  \par\normalfont\vspace{1em}
}

\begin{document}

\begingroup
\makeatletter
\let\raggedright\centering
\makeatother
\maketitle
\endgroup
\begingroup
\renewcommand{\thefootnote}{}
\footnotetext{Correspondence to Haozhe Chen at \href{mailto:hc5019@princeton.edu}{hc5019@princeton.edu} and Zhuang Liu at \href{mailto:zhuangl@princeton.edu}{zhuangl@princeton.edu}.}
\endgroup

% Links
\vspace{-0.8\baselineskip}
\begin{center}
    \github \href{https://github.com/zlab-princeton/ceobench-src/tree/main}{\textbf{Code}} \quad
    \worldwideweb \href{https://ceobench.com}{\textbf{Blog}} \quad
    \trajectoryicon \href{https://ceobench.com/trajectory-viewer/}{\textbf{Trajectory}}
\end{center}

% Abstract
\newcommand{\abstractcontent}{%
Language model agents are becoming proficient executors at isolated, short-horizon tasks such as software engineering and customer service. Yet real-world challenges require a combination of sophisticated skills that remain largely untested in agents: (1) navigating long horizons amid uncertainty; (2) acquiring information in noisy environments; (3) adapting to a changing world; (4) orchestrating multiple moving parts toward a coherent goal. We introduce \textsc{CEO-Bench}, which evaluates these capabilities together by simulating a representative real-world task: operating a startup for 500 days. An agent manages pricing, marketing, budgeting, and many other aspects of a fictional company through a programmable Python interface, operating in the same environment and facing the same challenges as a human CEO. Success demands analyzing noisy, interconnected business databases, translating signals into sound strategy, and coordinating many decisions with programming. The strongest agents write sophisticated code that forecasts churn regimes, billing timing, customer losses, and future cash under different scenarios. Even so, most state-of-the-art models struggle in this environment. Only Claude Fable 5, GPT-5.6 Sol, and Claude Opus 4.8 finish above the \$1M starting balance, and all evaluated models remain below the rule-based baseline. \textsc{CEO-Bench} takes a first step toward measuring the intelligence required to drive sustained, adaptive progress over time.
}

% Teaser above abstract

\begin{figure}[!ht]
  \centering
  \includegraphics[width=\textwidth]{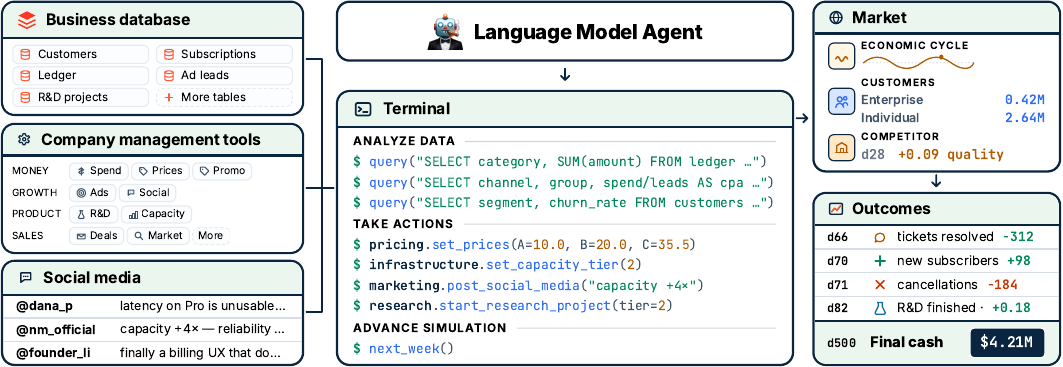}
  \caption{\textbf{\textsc{CEO-Bench} evaluates general long-horizon agent capabilities by simulating a startup over 500 days in a realistic and challenging environment.} The agent operates through a programmable interface with access to business databases, company management tools, and social media. Outcomes are driven by a partially observable, noisy, and evolving market with delayed and coupled consequences.}
  \label{fig:teaser}
\end{figure}

\vspace{0.3cm}

\begin{abstractblock}
\abstractcontent
\end{abstractblock}

\newpage
\begin{figure}[t]
  \centering
  \includegraphics[width=\textwidth]{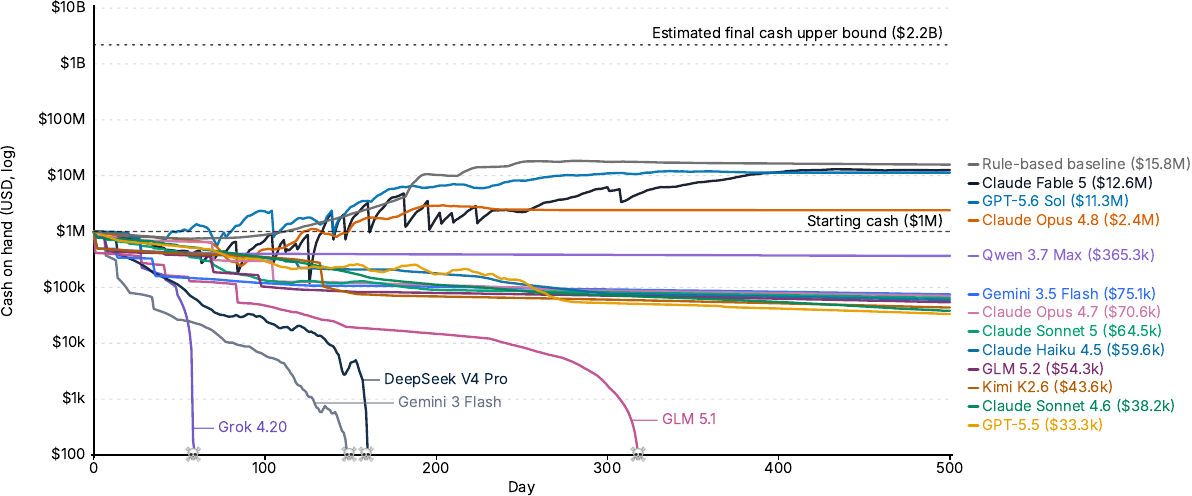}
  \caption{\textbf{Cash on hand over time for each model's best run}. Most state-of-the-art models struggle to complete the simulation without bankruptcy. Claude Fable 5, GPT-5.6 Sol, and Claude Opus 4.8 grow cash above the \$1M starting balance for their best runs, while all evaluated models remain below the rule-based baseline. Current models still struggle to combine long-horizon planning, noisy information gathering, adaptation, and coordinated execution over time.}
  \label{fig:cash-survival}
\end{figure}
\section{Introduction}

Language model agents are becoming increasingly capable at short-horizon tasks.
They can fix a GitHub issue~\citep{jimenez2024swebench}, follow a service policy in dialogue~\citep{yao2024taubench}, or complete a web workflow~\citep{zhou2024webarena}.
These are real skills, but they share a simple shape: the agent gets a clear goal, acts for a short time, and receives feedback quickly.
As agents approach reliable execution of such individual tasks, a natural next question is what we should expect them to do after the local task is no longer the bottleneck. 

Human intelligence goes beyond local execution~\citep{newell1972human,simon1955behavioral}. Many consequential human achievements are not single well-specified tasks, but long chains of decisions made under uncertainty: choosing what to prioritize, allocating limited resources, interpreting noisy signals, and adapting as conditions change~\citep{simon1955behavioral,march1991exploration,teece1997dynamic}. Future agents will need the same kind of sustained strategic control if they are to move beyond task completion and operate effectively in the real world.

Early agent evaluations such as SWE-bench~\citep{jimenez2024swebench}, WebArena~\citep{zhou2024webarena}, and $\tau$-bench~\citep{yao2024taubench} evaluate real-world skills, but they are scoped to short episodes with quickly observed outcomes.
GDPval~\citep{patwardhan2025gdpval} broadens evaluation to economically valuable work, but remains a one-shot deliverable rather than a persistent process. Agentic-memory benchmarks test agents' ability to use information over time, but they primarily measure storage and retrieval skills~\citep{hu2025memoryagentbench,he2026memoryarena}. Vending-Bench~\citep{backlund2025vendingbench,backlund2025vendingbench2} and Accounting-Bench~\citep{penrose2025accountingbench} take a first step toward evaluating agents in long-horizon simulated environments. Yet these settings involve a narrow set of decisions and largely stable environments. They do not test whether agents can coordinate many interdependent actions, acquire information from noisy feedback, and devise strategy amid delayed consequences and changing conditions.

% Agentic-memory benchmarks move closer by testing whether agents can retain and use information across interactions. MemoryAgentBench emphasizes storing and recalling evolving user information, while MemoryArena adds multi-session tasks where earlier experience can affect later actions~\citep{hu2025memoryagentbench,he2026memoryarena}. However, their focus remains on whether agents preserve useful information, not whether they can turn accumulated evidence into productive signals for steering a complex, changing system. Long-horizon operation benchmarks take another step by evaluating agents over extended processes, such as running a vending-machine business or closing monthly books from software company data~\citep{backlund2025vendingbench, backlund2025vendingbench2,penrose2025accountingbench}. Yet these settings still involve a limited set of decisions and largely stable environments. They also do not test whether agents can coordinate many interdependent choices under hidden state, noisy feedback, delayed consequences, and changing conditions.

The next stage of agent evaluation calls for a shift toward settings where \textit{actions accumulate over long horizons; environment state is only observable through indirect evidence; feedback is noisy and delayed; and conditions continue to change}. Success depends on more than individual capabilities. Agents must integrate them into coherent behavior, make long-horizon plans, turn accumulated evidence into actionable signals, coordinate interdependent decisions, and continuously adapt strategy as new information arrives.
\begin{figure}[t]
  \centering
  \includegraphics[width=\textwidth]{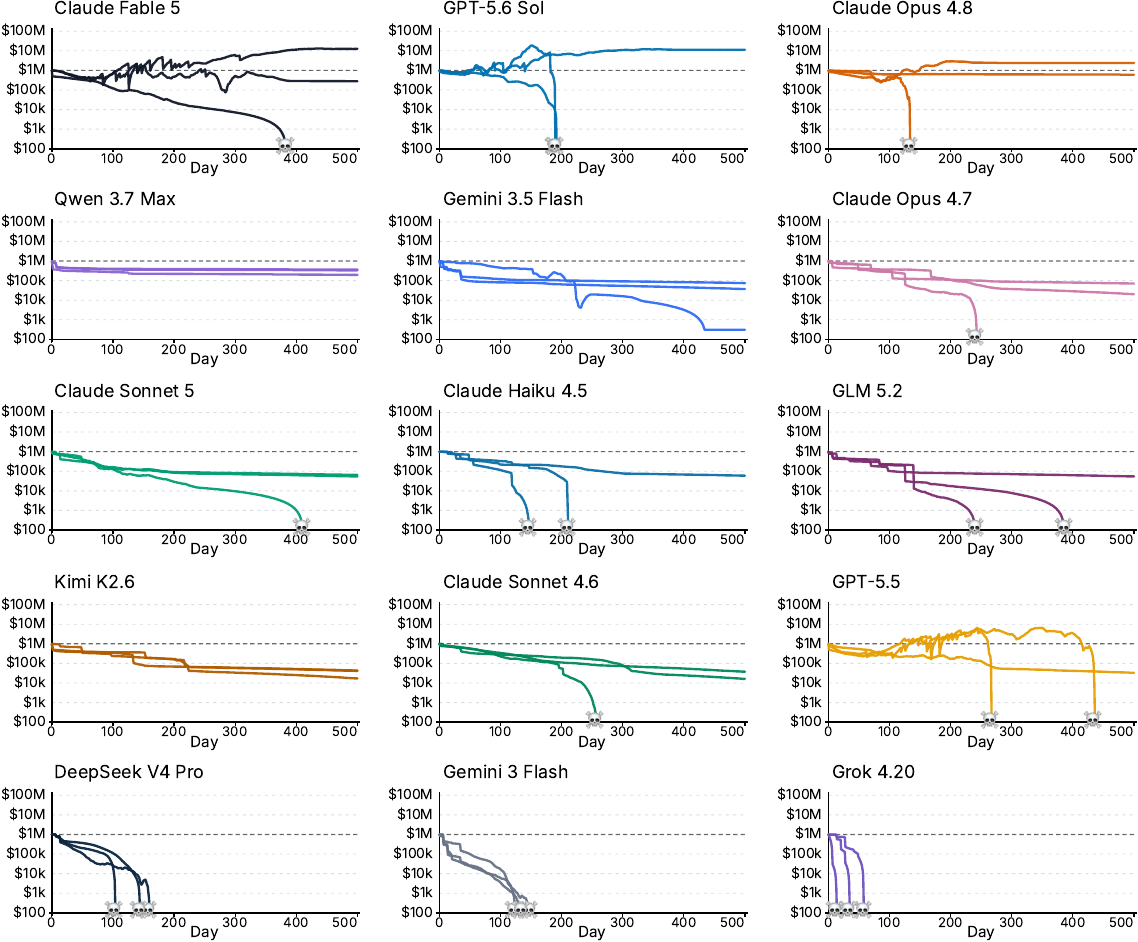}
  \caption{\textbf{Cash on hand over time for each of the three runs per model.} We show all experiments for each model to describe full behavior patterns. We also release all agent action trajectories in an \href{https://ceobench.com/trajectory-viewer/}{interactive trajectory viewer}.} 
  \label{fig:cash-trajectories-grid}
\end{figure}
\FloatBarrier

\begin{wrapfigure}[20]{r}{0.50\textwidth}
  \centering
  \includegraphics[width=\linewidth]{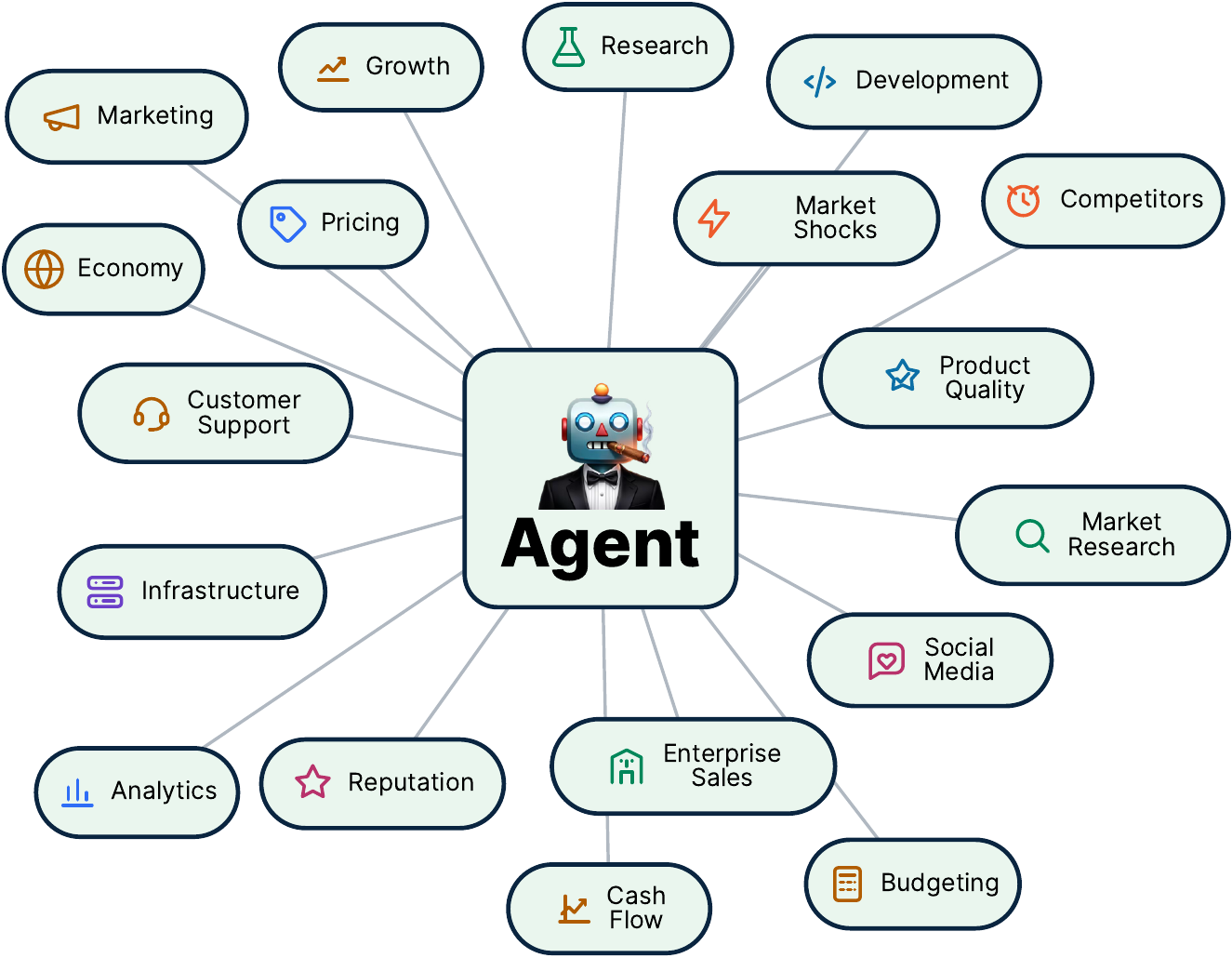}
  \caption{\textbf{Running a startup requires coordinating many moving parts,} making it a fitting choice as a canonical task evaluating agent's skills to steer complex decisions across long-horizon.}
  \label{fig:agent-complexity}
\end{wrapfigure}

\textsc{CEO-Bench} instantiates this challenge in a realistic, large-scale startup simulation, where an agent runs a company for 500 days through a programmable Python interface with 34 tools and a 19-table business database. We show the structure of \textsc{CEO-Bench} in Fig.~\ref{fig:teaser}. Beyond issuing individual tool calls, the agent writes and executes code, querying the database with SQL to analyze the company's state and composing the available tools into custom workflows. It thus operates in the same environment and faces the same challenges as a human running the company, and the task demands coding and data-analysis skill together with strategic thinking. As shown in Fig. \ref{fig:agent-complexity}, the agent must coordinate diverse operating decisions across pricing, growth, product, operations, communication, enterprise sales, and more. Decisions play out over realistic business timelines: revenue arrives on billing cycles, R\&D takes days to weeks, and mistakes surface later through churn and reputation, forcing long-horizon reasoning under uncertainty. Much of the state is hidden, so the agent must infer customer satisfaction, willingness to pay, and shifting preferences from noisy signals in data analytics and social media. At the same time, we design the environment to keep changing as a result of customer preference drift, macroeconomic cycles, and competitor shocks. Success requires continually revising strategy in response to these shifting conditions while maintaining coherent decisions across the business.

Our evaluation shows that this challenge remains difficult for agents built on current state-of-the-art models. We show in Fig.~\ref{fig:cash-survival} that while most agents can produce valid tool calls and analytics queries, they struggle to sustain coherent strategy over time and often bankrupt before completing the simulation. Claude Fable 5, GPT-5.6 Sol, and Claude Opus 4.8 finish above the \$1M starting balance, and no evaluated model reaches the non-LLM rule-based baseline. We show in Fig. \ref{fig:cash-trajectories-grid} that even the strongest models fail to consistently profit across experiments. 
Our analysis shows that performance correlates with core capabilities targeted by \textsc{CEO-Bench}: inferring hidden structure from noisy data, forecasting delayed consequences, and adapting to competitive pressure. By inspecting agent action trajectories, we find distinctive behavior patterns across models. For example, Claude Fable 5 keeps adjusting its strategy late into the run, Claude Opus 4.8 explores broadly early and then becomes more passive after building a cash cushion, and Claude Opus 4.7 narrows earlier into spend cuts and cash preservation. Claude Opus 4.8 reaches a larger customer base initially and then drops to zero customers mid-simulation, while Claude Fable 5 sustains a smaller customer base throughout the simulation. These results show that \textsc{CEO-Bench} \textit{exposes fine-grained behavioral patterns} that remain invisible to existing evaluations, while \textit{revealing substantial headroom in models' ability} to integrate individual capabilities into coherent, adaptive behavior over extended horizons.

\begin{table}[t]
  \centering
  \tablefontsize
  \setlength{\tabcolsep}{3pt}
  \renewcommand{\arraystretch}{0.92}
  \begin{tabular}{>{\raggedright\arraybackslash}p{0.18\textwidth} >{\raggedright\arraybackslash}p{0.42\textwidth} >{\raggedright\arraybackslash}p{0.33\textwidth}}
    \toprule
    \textbf{Category} & \textbf{Actions} & \textbf{Example tools} \\
    \midrule
    \textbf{Database query} & Query 19 business SQL databases and conduct data analytics & \texttt{query} \\
    \midrule
    \textbf{Monetization} & Set prices, usage quotas, discounts, and in-product ads & \texttt{pricing.set\_prices}, \texttt{pricing.set\_usage\_quotas} \\
    \midrule
    \textbf{Growth and market expansion} & Allocate targeted advertising spend and promotion across channels and customer groups & \texttt{marketing.set\_targeted\_ad\_spend}, \texttt{marketing.set\_lead\_promotion} \\
    \midrule
    \textbf{Product quality and R\&D} & Choose model tiers, fund day-to-day development, and launch research projects & \texttt{pricing.set\_model\_tiers}, \texttt{research.start\_research\_project} \\
    \midrule
    \textbf{Reliability} & Buy infrastructure capacity and fund customer support & \texttt{infrastructure.set\_capacity\_tier}, \texttt{analytics.set\_targeted\_ops\_spend} \\
    \midrule
    \textbf{Enterprise sales} & Conduct multi-turn negotiations over price and plan with enterprise prospects and renewals & \texttt{enterprise.send\_enterprise\_deal}, \texttt{enterprise.reject\_enterprise\_deal} \\
    \midrule
    \textbf{Information acquisition} & Pay for market research to discover new customer groups and learn more about existing groups & \texttt{market.research\_market}, \texttt{market.research\_group} \\
    \midrule
    \textbf{Public communication} & Monitor social media for customer complaints, competitor news, and economic trends, then post or reply to influence growth & \texttt{marketing.post\_social\_media}, \texttt{analytics.get\_social\_posts} \\
    \bottomrule
  \end{tabular}
  \caption{\textbf{Agent action space} categories and example tools in \textsc{CEO-Bench}. These tools enable agents to design diverse operation strategies but also pose challenges on coordinating many actions toward one coherent goal.} 
  \label{tab:action-space}
\end{table}

\section{Designing \textsc{CEO-Bench}}

In this section, we provide an overview of how the \textsc{CEO-Bench} simulator works. Then, we describe how we design the action interface to make the task open-ended. Finally, we detail world mechanics design considerations that make the task challenging and realistic.
\subsection{How \textsc{CEO-Bench} Works}
\label{sec:simulator}
In \textsc{CEO-Bench}, an agent runs a fictional subscription-software company called \emph{NovaMind} for 500 simulated days. It begins on day one with zero customers and \$1M in cash and is graded on cash on hand at the end. If cash ever falls strictly below zero, bankruptcy ends the simulation. We provide an overview of the simulator mechanics in this section and include full details in Appendix~\ref{app:simulator-details}.

\textbf{What an agent can do.} For each simulated week, the agent can take actions for unlimited turns across 34 tools in the categories displayed in Table~\ref{tab:action-space}. These categories cover pricing and plan design, growth and market expansion, product quality and research, reliability and support, information acquisition, public communication, and enterprise sales. Each tool accepts fine-grained structured arguments, so agents can compose a large space of possible policies. Section~\ref{sec:interface} explains the tool interface design in more detail.

\textbf{How an agent makes and loses money.} An agent makes profits through customer subscription payments and in-product ad monetization. We abstract the company product that customers subscribe to as a numerical product quality. Higher product quality results in more product subscriptions and payments, but maintaining quality via development, research, infrastructure capacity, support, and model tier choices requires spending. Acquiring customers through advertising channels also costs money. Cash therefore changes through both immediate costs and delayed revenue effects. We show the calculation of cash change between each day and decompose each contributing factor in \eqrefc{eq:daily-cash-change}. We fully explain the role and mechanism of each factor in Appendix ~\ref{app:product-quality-usage-monetization}, \ref{app:satisfaction-retention-support} and~\ref{app:costs-cash-flow}.
\begin{equation}
\begin{adjustbox}{max width=\linewidth}
\(\displaystyle
\begin{aligned}
\underbrace{B_{t+1}-B_t}_{\substack{\text{daily cash}\\\text{change}}}
={}&
\underbrace{Y_t^{\mathrm{sub}}}_{\substack{\text{subscription}\\\text{payments}}}
+\underbrace{\sum_iY_{i,t}^{\mathrm{ads}}}_{\substack{\text{in-product}\\\text{ads}}}
-\underbrace{K_{\kappa_t}^{\mathrm{capacity}}}_{\substack{\text{capacity}\\\text{cost}}}
-\underbrace{\sum_p \chi_p^{\mathrm{usage}}U_{p,t}^{\mathrm{use}}}_{\substack{\text{usage compute}\\\text{cost}}}
-\underbrace{x_t^{\mathrm{ops}}}_{\substack{\text{support}\\\text{spending}}}
-\underbrace{x_t^{\mathrm{dev}}}_{\substack{\text{dev}\\\text{spending}}}\\
&-\underbrace{X_t^{\mathrm{target\text{-}ops}}}_{\substack{\text{targeted}\\\text{support}}}
-\underbrace{\sum_gx_{g,t}^{\mathrm{target\text{-}dev}}}_{\substack{\text{targeted}\\\text{dev}}}
-\underbrace{\sum_{c,g}x_{c,g,t}^{\mathrm{ads}}}_{\substack{\text{acquisition}\\\text{ads}}}
-\underbrace{N_t^{\mathrm{lead}}c^{\mathrm{lead}}}_{\substack{\text{lead}\\\text{acquisition}}}
-\underbrace{K_t^{\mathrm{market}}}_{\substack{\text{market}\\\text{research}}}
-\underbrace{K_t^{\mathrm{group}}}_{\substack{\text{group}\\\text{research}}}
-\underbrace{K_t^{\mathrm{project}}}_{\substack{\text{research}\\\text{projects}}}
\end{aligned}
\)
\end{adjustbox}
\label{eq:daily-cash-change}
\end{equation}

\textbf{Modeling customers and indirect feedback.} There are 26 customer groups in the simulator. Each customer group consists of a distribution of hidden price and quality preferences, such as a maximum willingness to pay and a minimum accepted quality at each price. Each customer is created by sampling its unique preference parameters from a group distribution. At a subscription plan's price, a customer subscribes if the offered product quality exceeds the customer's minimum accepted quality. The customer may switch plans if another plan gives a better quality surplus and may cancel if no plan remains acceptable. We show the mathematical definition of a customer's price-quality preference curve in \eqrefc{eq:customer-participation-curve}. We describe full details of each factor in Appendix~\ref{app:simulator-details}, Subsection~\ref{app:customers-plans-participation}. Example curve plots appear in Fig.~\ref{fig:customer-participation-curves}. Customer satisfaction changes company reputation, and reputation affects the new customer acquisition rate. The agent does not directly observe satisfaction, willingness to pay, or quality thresholds. It instead infers feedback by analyzing subscription, churn, support, revenue, and reputation data and by monitoring simulated social media.

\textbf{Customer acquisition and enterprise negotiation.} Agents acquire new customers by spending on advertising channels. Each customer group reacts differently to each ad channel, so the same spend can produce different acquisition rates across groups. Reputation, social media reactions, market saturation, demand surges, and macroeconomic conditions also affect acquisition speed. We show the calculation of expected new prospective customers for group $g$ on day $t$ in \eqrefc{eq:expected-prospective-customers}. We describe full details of each factor in Appendix~\ref{app:reputation-social-acquisition} and~\ref{app:market-discovery-nonstationarity}. We sample daily from a Poisson distribution parameterized by this expectation. Market research can reveal additional customer groups and improve what the agent knows about known groups. Enterprise customers follow the same price and quality logic, but deals are negotiated through offers, counter-offers, reply delays, and possible rejection.
\begin{equation}
\begin{adjustbox}{max width=\linewidth}
\(\displaystyle
\begin{aligned}
\underbrace{\mathbb{E}\!\left[n_{g,t}^{\mathrm{prospect}}\right]}_{\substack{\text{expected new prospective}\\\text{customers for group }g}}
={}&
\underbrace{R_{g,t}}_{\substack{\text{reputation}\\\text{in group }g}}
\cdot
\underbrace{D_{g,t}}_{\substack{\text{market saturation}\\\text{for group }g}}
\cdot
\underbrace{C_t}_{\substack{\text{calendar}\\\text{cycle}}}
\cdot
\underbrace{M_{g,t}}_{\substack{\text{macro econ}\\\text{cycle}}}
\cdot
\underbrace{A_{g,t}}_{\substack{\text{social media}\\\text{reaction}}}
\cdot
\underbrace{Z_t}_{\substack{\text{demand}\\\text{surge}}}
\cdot\left(
\underbrace{\sum_c\frac{x_{c,g,t}L_{c,g,t}}{x_{\mathrm{ad}}}}_{\substack{\text{leads from each}\\\text{ad channel}}}
+\underbrace{\sum_hN_{h,t}W^{\mathrm{net}}_{h,g}}_{\substack{\text{networking effect}\\\text{from each group}}}
\right)
\end{aligned}
\)
\end{adjustbox}
\label{eq:expected-prospective-customers}
\end{equation}
\textbf{Product quality and competitor pressure.} Product quality is affected by daily development, research projects, model tier choices, targeted development, infrastructure capacity, support spending, usage quotas, and in-app ad strength. These controls shape customer experience through base product quality, quota fulfillment, system overload, support delays, relationship history, and ad load. Quota shortfalls multiply the whole perceived-quality expression. Competitors add pressure by periodically raising customer quality expectations. Broad product development and research can make competitors catch up faster, while targeted development for specific groups is harder to copy and lets competitors catch up more slowly. We show the computation of a customer's perceived product quality and breakdown of each factor in \eqrefc{eq:perceived-quality-decomposition}. We describe full details of each factor in Appendix~\ref{app:product-quality-usage-monetization}, \ref{app:satisfaction-retention-support}, and~\ref{app:market-discovery-nonstationarity}.
\begin{equation}
\begin{adjustbox}{max width=\linewidth}
\(\displaystyle
\begin{aligned}
\underbrace{Q_{i,t}^{\mathrm{perc}}}_{\substack{\text{quality perceived}\\\text{by customer }i}}
={}&
\underbrace{\phi_{i,p,t}^{\mathrm{quota}}}_{\text{quota factor}}
\Bigg[
\underbrace{m_p}_{\text{model-tier effect}}\left(
\underbrace{q_0}_{\text{initial quality}}
+\underbrace{b_t^{\mathrm{shared}}}_{\text{dev improvement}}
+\underbrace{b_{g,t}^{\mathrm{group}}}_{\text{targeted dev improvement}}
\right)
-\underbrace{\beta_o o_t}_{\text{overload penalty}}\\
&+\underbrace{\beta_r(r_{i,t}-r_0)}_{\text{customer relationship}}
+\underbrace{\beta_d\log\!\left(\alpha_d+d_{i,t}/d_0\right)}_{\text{customer stickiness}}
-\underbrace{\beta_I I_{i,t}}_{\text{open issues penalty}}
-\underbrace{\eta_i^{\mathrm{ads}}a_{i,t}^{\mathrm{eff}}}_{\text{in-app ads penalty}}
\Bigg]
\end{aligned}
\)
\end{adjustbox}
\label{eq:perceived-quality-decomposition}
\end{equation}

\textbf{Changing world imposes challenges.} The world evolves over time through macroeconomic trends, interconnected reputation propagation, market saturation, demand surges, and competitor pressure. These factors affect acquisition, retention, and enterprise deal outcomes. The challenge is that the agent observes only partial and delayed evidence of these changes. It must infer hidden customer and market conditions from traces, choose actions whose effects arrive on different time scales, and revise its policy as the company and market move. Section~\ref{sec:interface} explains the interface design, and Section~\ref{sec:world} describes the design principles that make the simulator mechanics realistic and challenging.
\begin{figure}[t]
  \centering
  \includegraphics[width=\textwidth]{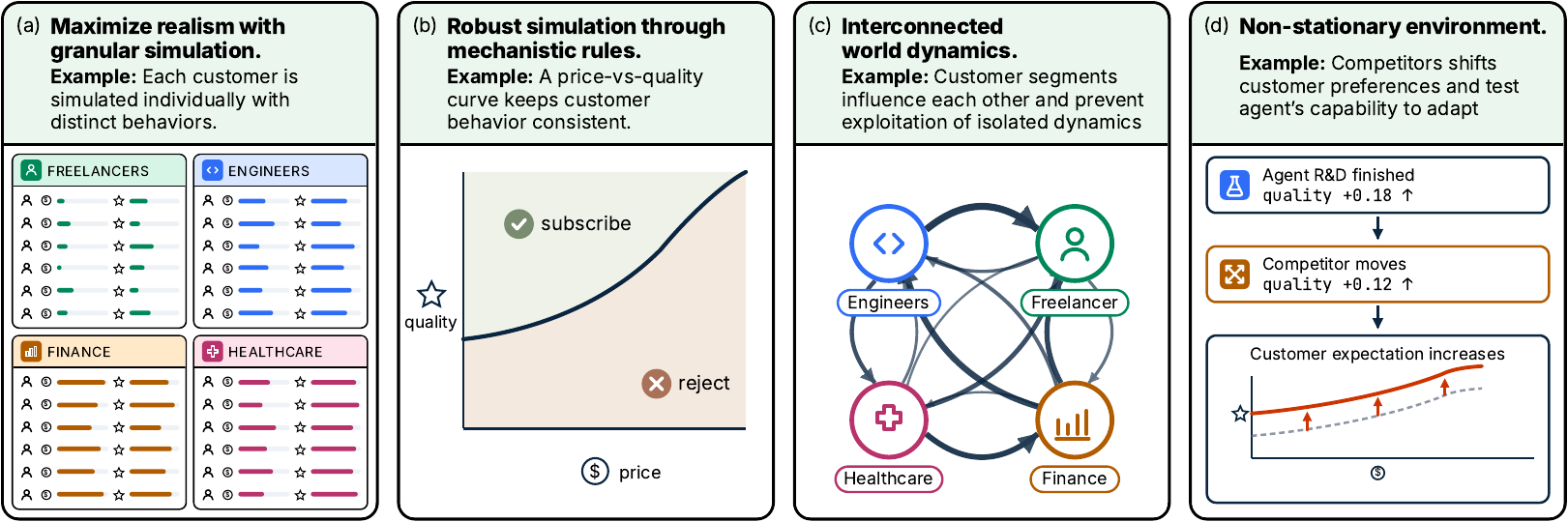}
  \caption{\textbf{Major design principles} behind \textsc{CEO-Bench}'s world mechanics and example designs that follow the principles.}
  \label{fig:world-mechanics}
\end{figure}

% Costs are paid daily; subscription revenue arrives on monthly billing cycles; R\&D outcomes return weeks after they are commissioned; reputational damage may take a month to surface in churn; the macroeconomic backdrop is visible only with a multi-week lag.

% The world the agent operates in is opaque.
% Customer satisfaction, group-level reputation, churn probabilities, and the drifting preference parameters that govern demand are never directly observable.
% Customers leave but do not explain why.
% Competitors act on schedules the agent cannot see.
% The only way to make sense of what is happening is to query the operational database, read social-media posts, commission surveys, and reason from indirect evidence---the same indirect evidence a real operator works from.

\subsection{How We Make \textsc{CEO-Bench} Rigorous and Challenging}

\label{sec:world}

We design \textsc{CEO-Bench}'s world mechanics to be an expressive emulation of the real world, while remaining mechanistic so that success depends on genuine skills rather than exploiting brittle simulations. We describe seven core principles in our world mechanics design below and illustrate four examples in Fig.~\ref{fig:world-mechanics}.

\paragraph{Maximize realism with granular simulation.}
The simulator models 26 customer groups and individual customers within each group rather than only aggregate demand. Each customer has its own acquisition path, subscription state, price exposure, usage, satisfaction, and cancellation trajectory. Customers are also organized into diverse groups with different needs, budgets, price sensitivities, ad channel effectiveness, support expectations, and behavioral patterns. This granularity increases the complexity of world dynamics and widens the set of viable strategies.

\paragraph{Robust simulation with mechanistic rules.}
The world emulates real business behavior while maintaining stable cause-and-effect relationships. Almost all simulator outcomes are generated by explicit mechanisms rather than by using an LLM as an opaque judge. For example, customers decide whether to subscribe by comparing product value against price through a microeconomics-motivated participation rule~\citep{mussa1978monopoly}. This design aims to avoid failure modes in benchmarks such as Vending-Bench~\citep{backlund2025vendingbench, backlund2025vendingbench2}, where an LLM-simulated supplier can reward agent's unrealistic verbal promises.

\paragraph{Consistent simulation under stochasticity.}
While we inject stochasticity into world dynamics to emulate real-world noise, we maintain consistency across runs with independent random number generators for different simulator components. For example, under the same random seed, after calling the market research tool multiple times, the agent always discovers the same sequence of new market groups, independent of actions in other areas.

\paragraph{Hidden information and indirect feedback.}
\textsc{CEO-Bench} tests whether agents can gather information in a partially observable world. The agent receives only information that a real start-up manager could plausibly observe: dashboards, database records, social-media posts, research reports, and negotiation history. It does not observe true customer satisfaction, latent willingness to pay, churn propensity, competitor schedules, or demand parameters. Instead, it must infer these hidden variables indirectly, for example, by gauging customer satisfaction and complaints through social media or detecting competitor moves by analyzing cancellation behavior.

\paragraph{Interconnected world dynamics.}
We design the simulated world to make it difficult to isolate a single causal relationship and hill-climb on it. Every decision can influence many other parts of the market. For example, reputation propagates across related groups, so a quality failure in one enterprise group can spill into nearby groups and eventually affect consumer demand. Increasing satisfaction of influential customer groups can boost growth more effectively than ads.

\paragraph{Delayed and uncertain consequences.}
Many actions have delayed and uncertain effects, forcing long-horizon decision making under uncertainty. Costs may appear immediately, while the corresponding revenue, retention, research, or reputation effects arrive weeks later. R\&D projects have stochastic completion timelines and quality improvements, so investing more does not deterministically produce an immediate gain. Enterprise negotiations also unfold over stochastic delays, making it costly to wait too long but risky to overreact to any single turn. We show the types of distributions used and example usage in Table~\ref{tab:stochastic-mechanisms}.
\begin{table}[!t]
  \centering
  \tablefontsize
  \setlength{\tabcolsep}{4pt}
  \renewcommand{\arraystretch}{0.96}
  \begin{tabular}{>{\raggedright\arraybackslash}p{0.22\textwidth} >{\raggedright\arraybackslash}p{0.35\textwidth} >{\raggedright\arraybackslash}p{0.33\textwidth}}
    \toprule
    \textbf{Distribution} & \textbf{Example use in simulator} & \textbf{Motivation} \\
    \midrule
    \textbf{Normal} & R\&D project quality gain & Captures uncertain payoff \\
    \midrule
    \textbf{Poisson} & Daily new prospective customers for a group & Models rate-based counts \\
    \midrule
    \textbf{Bernoulli} & Involuntary cancellation event & Models binary shocks \\
    \midrule
    \textbf{Uniform} & Reputation damage noise & Adds bounded uncertainty \\
    \midrule
    \textbf{Log-normal} & Competitor quality-jump magnitude & Models skewed positive shocks \\
    \bottomrule
  \end{tabular}
  \caption{\textbf{Stochastic mechanisms in \textsc{CEO-Bench}.} The simulator uses a variety of stochastic variables to model real-world uncertainties.}
  \label{tab:stochastic-mechanisms}
\end{table}

\paragraph{Non-stationary environment.}
Agents must continually gather new information and adapt because the environment changes over the course of a simulation. Competitors place adaptive pressure on product quality. Customer behavior also drifts over time, with different groups shifting at different rates in price sensitivity and quality expectations. Macroeconomic trends add another changing background process, affecting willingness to pay and enterprise seat counts across expansions and contractions.

\subsection{A Versatile Action Interface Between World and Agent}
\label{sec:interface}

\begin{figure}[!t]
  \centering
  \includegraphics[width=\textwidth]{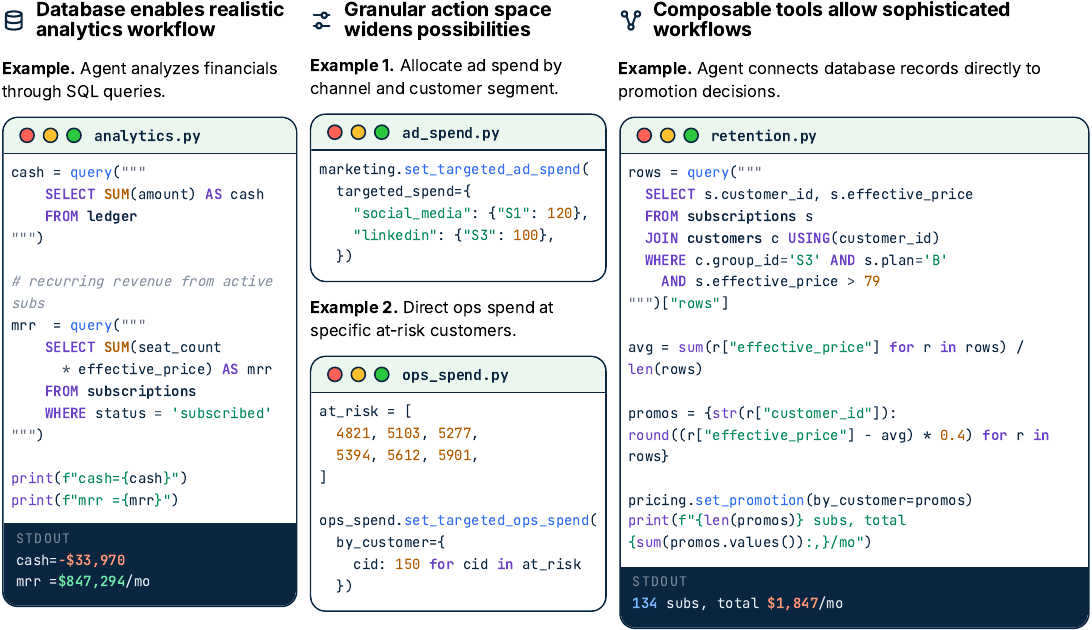}
  \caption{\textbf{Agents interact with \textsc{CEO-Bench} through a versatile Python interface.} Left: We give the agent access to diverse business databases to test its information acquisition capability through a realistic data analytics workflow. Middle: We widen agents' opportunity space by enabling them to take fine-grained actions. Right: This interface design allows the agent to compose tools into sophisticated custom workflows.}
  \label{fig:interface-examples}
\end{figure}
We design a programmable tool interface, so agents can effectively manage granular action spaces and organize them into custom workflows.

\paragraph{Composable action interface in Python.}
Terminal-based computer-use agents have become a general form factor across tasks~\citep{anthropic2026claudecode,openai2026codexcli,opencode2026,pi2026}. We make evaluating \textsc{CEO-Bench} easy with any of these agents by exposing the action surface to the agent via a Python package, \verb|novamind_api|. An agent manages the company by calling functions in \verb|novamind_api| in a Python script and executing the script in its terminal.
This design maximizes flexibility for an agent to build its own infrastructure on top of the API. In Fig.~\ref{fig:interface-examples} (right), we show an example where, rather than calling a tool once per customer, an agent connects to the database via its custom data-driven promotion management system and applies promotion decisions efficiently at scale.

\paragraph{Granular action spaces.}
We allow agents to act at fine granularity to create a rich space of strategic tradeoffs, failure modes, and opportunities for adaptation. Although the interface contains a finite set of tools, each tool accepts fine-grained structured arguments, so agents can compose a combinatorially large space of possible actions. In Fig.~\ref{fig:interface-examples} (middle), we show examples where the agent allocates advertising spend by (ad channel, customer group) pair and decides operations spending on individual customers.

\paragraph{Large-scale and realistic databases.}
We give the agent access to a 19-table operational database covering orders, contracts, subscriptions, the cash ledger, the social-media feed, configuration history, ad-channel attribution, and support tickets, among others.
The schema mirrors what a real software company's analytics stack would expose, testing the agent's capability to gather information via an analytics workflow that resembles real-world software company operations. In Fig.~\ref{fig:interface-examples} (left), we show an example where the agent analyzes its revenue through database queries.

\paragraph{Social media.}
The agent can read a simulated public feed of customer complaints, competitor announcements, and macroeconomic trends. Agents can also reply and post on social media. Reactions to the agent's posts on social media can also influence the rate of new customer acquisition. We test the agent's capability to both perceive and act in a chaotic natural-language domain.

\section{Experiments and Results}
\begin{table}[!t]
  \centering
  \tablefontsize
  \setlength{\tabcolsep}{3.1pt}
  \renewcommand{\arraystretch}{0.84}
  \noindent\makebox[\textwidth][c]{%
  \begin{tabular}{l c r c c r r}
    \toprule
    \textbf{Model} & \textbf{Bankruptcy} & \makecell{\textbf{Best-run}\\\textbf{cash (\$)}} & \makecell{\textbf{Max survival}\\\textbf{days}} & \makecell{\textbf{Mean survival}\\\textbf{days $\pm$ std}} & \makecell{\textbf{Turns}\\\textbf{/week}} & \makecell{\textbf{Best run}\\\textbf{API cost}} \\
    \midrule
    Claude Fable 5     & 1/3 & 12{,}630{,}078 & 500 & $461.7 \pm  54.2$  &  9.86 & \$265.43 \\
    GPT-5.6 Sol        & 2/3 & 11{,}313{,}982 & 500 & $294.0 \pm 145.7$  & 25.96 & \$153.47 \\
    Claude Opus 4.8    & 1/3 &  2{,}399{,}209 & 500 & $378.0 \pm 172.5$  & 16.64 & \$348.49 \\
    Qwen 3.7 Max       & 0/3 &    365{,}346   & 500 & $500.0 \pm   0.0$  &  7.82 & --       \\
    Gemini 3.5 Flash   & 0/3 &     75{,}126   & 500 & $500.0 \pm   0.0$  & 58.37 & \$67.09  \\
    Claude Opus 4.7    & 1/3 &     70{,}620   & 500 & $414.3 \pm 121.2$  & 14.33 & \$92.32  \\
    Claude Sonnet 5    & 1/3 &     64{,}459   & 500 & $470.7 \pm  41.5$  & 20.50 & \$68.25  \\
    Claude Haiku 4.5   & 2/3 &     59{,}625   & 500 & $286.3 \pm 153.3$  & 19.35 & --       \\
    GLM 5.2            & 2/3 &     54{,}327   & 500 & $376.7 \pm 105.6$  & 25.24 & --       \\
    Kimi K2.6          & 0/3 &     43{,}598   & 500 & $500.0 \pm   0.0$  & 16.64 & --       \\
    Claude Sonnet 4.6  & 1/3 &     38{,}166   & 500 & $419.3 \pm 114.1$  & 14.37 & \$73.28  \\
    GPT-5.5            & 2/3 &     33{,}260   & 500 & $401.0 \pm  98.3$  & 30.41 & \$153.80 \\
    GLM 5.1            & 3/3 &          0     & 319 & $155.7 \pm 130.0$  & 98.66 & --       \\
    DeepSeek V4 Pro    & 3/3 &          0     & 160 & $136.3 \pm  23.7$  & 21.85 & --       \\
    Gemini 3 Flash     & 3/3 &          0     & 150 & $137.0 \pm  10.2$  & 15.81 & \$1.19   \\
    Grok 4.20          & 3/3 &          0     &  59 & $ 36.0 \pm  18.4$  & 13.49 & \$3.64   \\
    \midrule
    Rule-based baseline          & -- & 15{,}756{,}408 & 500 & --             & --   & --       \\
    Estimated final cash upper bound & -- & 2{,}200{,}000{,}000 & -- & --             & --   & --       \\
    \bottomrule
  \end{tabular}}
  \caption{\textbf{Benchmark results summary.} Most models fail to avoid bankruptcy or finish below the initial \$1,000,000 cash balance. Claude Fable 5, GPT-5.6 Sol, and Claude Opus 4.8 are the only evaluated models whose best runs finish above the starting balance, and all evaluated models remain below the rule-based baseline and the estimated upper bound of attainable final cash. \textsc{CEO-Bench} presents a challenging task for existing models.}
  \label{tab:results-overview}
\end{table}
\label{sec:experiments}
In this section, we describe our experiments and their results. We then conduct both qualitative and quantitative analysis to compare behaviors across models.
\subsection{Experimental Setup}
\label{sec:experimental-setup}

\textbf{Models.} We evaluate agents on the full 500-day \textsc{CEO-Bench} simulation. Each model is given \$1M starting cash. We run three simulations for each model with random seed 42. Each model uses the maximum reasoning effort provided. We evaluate Claude Fable 5, GPT-5.6 Sol, Claude Opus 4.8, Claude Opus 4.7, Claude Sonnet 5, Claude Sonnet 4.6, Claude Haiku 4.5, GPT-5.5, Qwen 3.7 Max, Gemini 3.5 Flash, Gemini 3 Flash, GLM 5.2, GLM 5.1, Kimi K2.6, DeepSeek V4 Pro, and Grok 4.20.

\textbf{Harness.} Terminal-based computer-use agents have become a general interface for automation: systems such as Claude Code, Codex, OpenCode, and Pi can perform diverse tasks and maintain memory by interacting with a terminal~\citep{anthropic2026claudecode,openai2026codexcli,opencode2026,pi2026}. We design \textsc{CEO-Bench} to be compatible with any such agent. To align the harness across all models, we implement a minimal terminal agent interface: we give each agent a Linux working directory and tools including bash, read-file, and edit-file. In early runs, we found that open-source harnesses such as OpenCode and Pi (pi-mono) did not manage context reliably enough for 500-day episodes, so our harness refreshes context by clearing action history and only keeping system prompt and an agent-editable memory file in context at the start of each simulated week.

\textbf{Result selection for analysis.} For results and analysis, we select the best run for each model as follows: (1) if at least one of the three runs avoids bankruptcy, we choose the run with maximum ending cash; (2) if all runs end in bankruptcy, we choose the run with the maximum number of simulation days before bankruptcy.

\subsection{Results Overview}
\label{sec:results-overview}
\begin{figure}[!t]
  \centering
  \includegraphics[width=\textwidth]{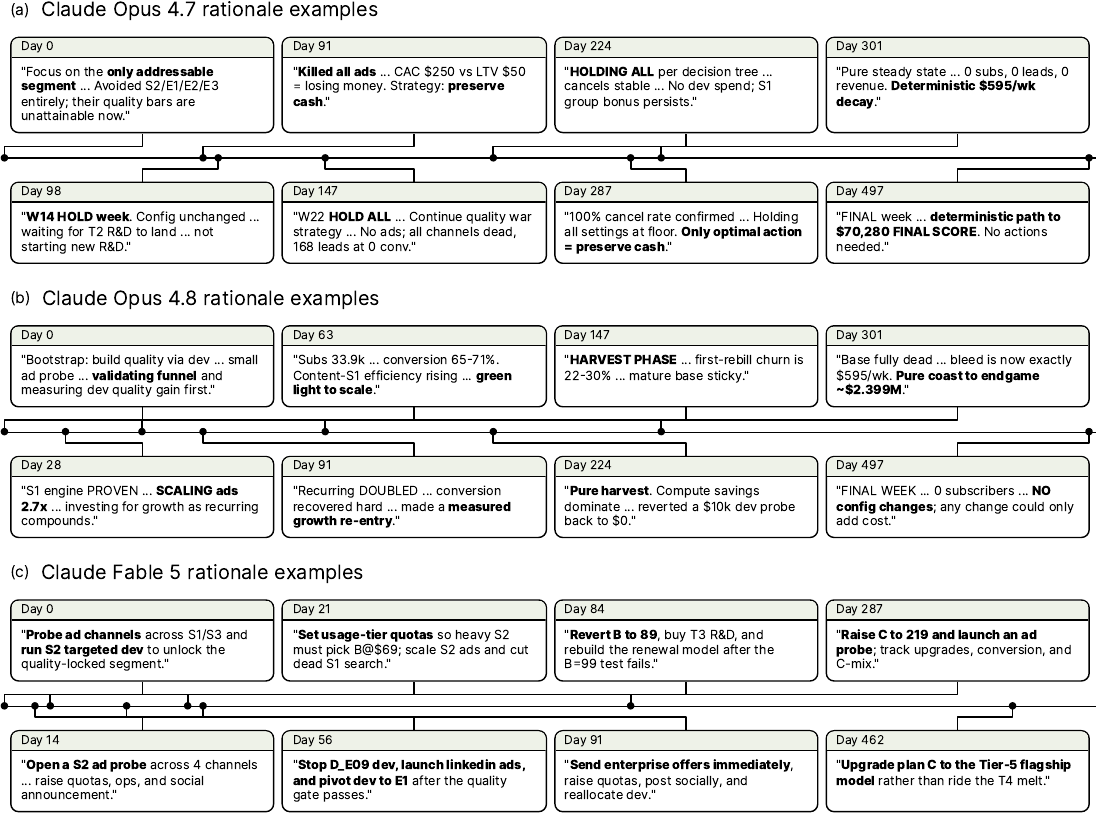}
  \caption{\textbf{Example memos written by Claude Opus 4.7, Claude Opus 4.8, and Claude Fable 5 in their workspaces during the best trajectory of each model.} Claude Fable 5 keeps adjusting its plan across the run, Claude Opus 4.8 explores broadly early and then becomes more passive, and Claude Opus 4.7 narrows earlier into waiting and protecting cash.}
  \label{fig:memo-timeline}
\end{figure}

\textbf{Overall results.} We show the best-run cash over time for each model in Fig.~\ref{fig:cash-survival}, and per-model trajectories across all three runs in Fig.~\ref{fig:cash-trajectories-grid}. We also show additional details in Table~\ref{tab:results-overview}. Most state-of-the-art models struggle to complete the simulation without bankruptcy. GPT-5.6 Sol finishes with \$11.31M in its best run, second only to Claude Fable 5, but its other two runs bankrupt around day 190. Claude Fable 5, GPT-5.6 Sol, and Claude Opus 4.8 are the only evaluated models whose best runs finish above the \$1M starting balance. GPT-5.5 reaches \$6.58M mid-run but is unable to preserve profitability across complete runs, ending at \$33K in its best completed run. Several models survive with positive cash below the starting balance, while GLM 5.1, DeepSeek V4 Pro, Gemini 3 Flash, and Grok 4.20 bankrupt on all runs. This preliminary evaluation shows that the strongest models demonstrate high-upside strategic behavior, but most models still fail to coordinate growth, quality, and cash flow over the full horizon.

\textbf{Rule-based baseline.} We include a simple rule-based heuristic baseline that uses no language-model calls during policy execution: it fixes prices, quotas, and model tiers, concentrates acquisition and targeted development on a small set of customer groups, and adjusts capacity from recent usage. We conduct a preliminary grid search over this rule template and display the best strategy in Fig.~\ref{fig:cash-survival} and Table~\ref{tab:results-overview}. The heuristic achieves a significant positive cash balance of \$15.76M, above all evaluated model agents. We show full details of this baseline strategy's design and configuration search in Appendix~\ref{app:rule-based-baseline}.

\textbf{Benchmark is far from saturated.} We estimate loosely the upper bound of achievable final cash be around \$2.2B. The estimation sums revenue from all 26 customer groups under maximum supportable pricing and subtracts the required costs for compute, capacity, development, operations, advertising, research, and acquisition. To obtain a conservative estimate, we further adjust downward for execution frictions, including issue-driven churn, enterprise negotiation friction, and acquisition delays. The resulting estimate remains far above the best observed model performance, indicating that \textsc{CEO-Bench} is far from saturated. We detail our estimation process in Appendix~\ref{app:upper-bound-estimation}. We release agent trajectories of all our experiments in an \href{https://ceobench.com/trajectory-viewer/}{interactive trajectory viewer}.

\subsection{A Look into Agent Behaviors}

In this section, we take a preliminary exploration in agent behaviors and compare them across models.

\begin{figure}[!t]
  \centering
  \includegraphics[width=\textwidth,trim=0 30pt 0 0,clip]{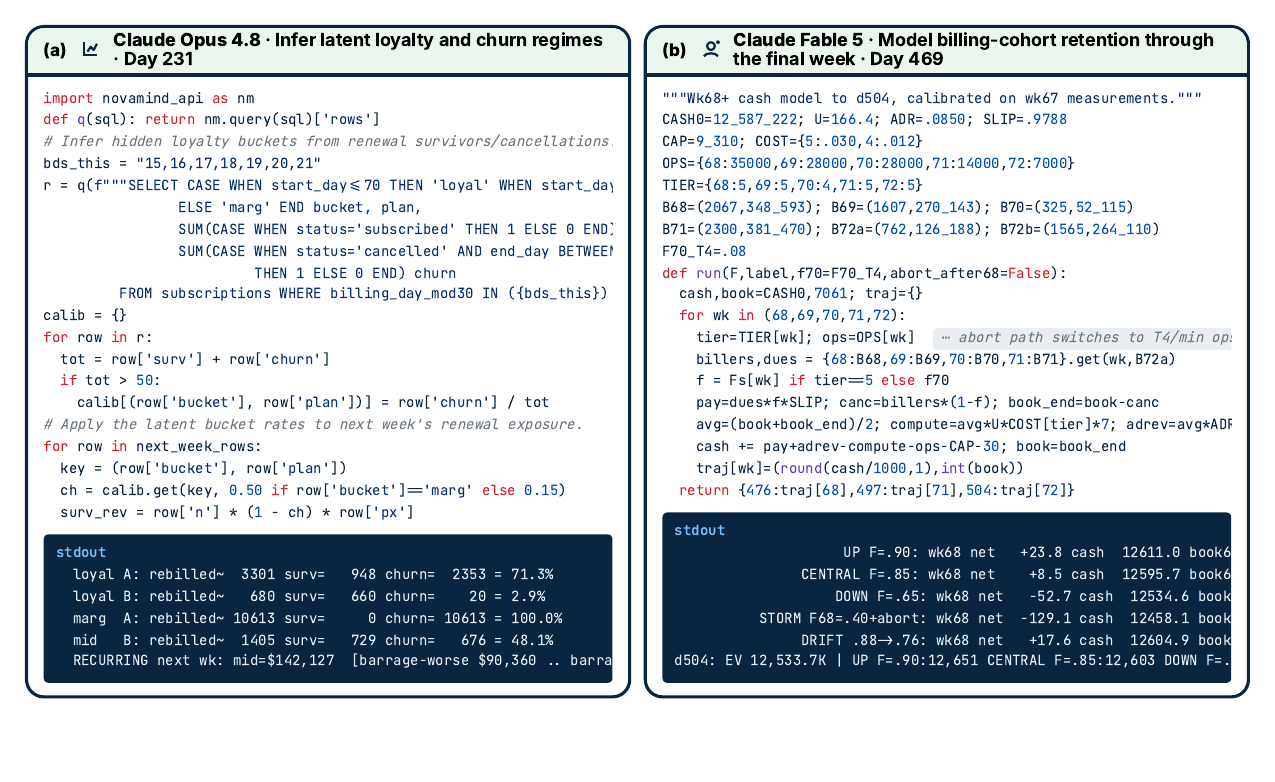}
  \caption{\textbf{Example code files written by top-performing agents during their best trajectories.} (a) Claude Opus 4.8 uses customer counts and cancellations to estimate future cash. (b) Claude Fable 5 tracks payment timing, customer losses, and final-week cash under different scenarios.}
  \label{fig:agent-code-example}
\end{figure}

\begin{figure}[!t]
  \centering
  \includegraphics[width=\textwidth]{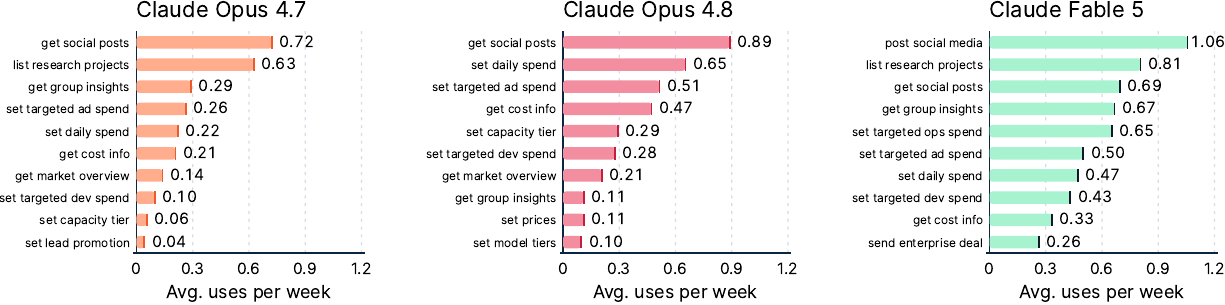}
  \caption{\textbf{Average per-week tool usage frequency for the best runs of Claude Opus 4.7, Claude Opus 4.8, and Claude Fable 5 (top 10 tools per model).} Claude Fable 5 and Claude Opus 4.8 show a more even spread across tools.}
  \label{fig:tool-usage}
\end{figure}

\textbf{Strong models explore wider strategy space.} In Fig.~\ref{fig:memo-timeline}, we show example memos written by agents over time. Claude Fable 5 keeps trying different ways to run the company as conditions change, and continues adjusting late in the run. Claude Opus 4.8 explores broadly at first, then becomes more passive after building a cash cushion. In contrast, Claude Opus 4.7 narrows much earlier, repeatedly choosing to spend less, wait, and protect cash. In Fig.~\ref{fig:tool-usage}, we find the same pattern quantitatively: Claude Fable 5 and Claude Opus 4.8 spread their actions more evenly across tools than Claude Opus 4.7.

\begin{figure}[!t]
  \centering
  \includegraphics[width=\textwidth]{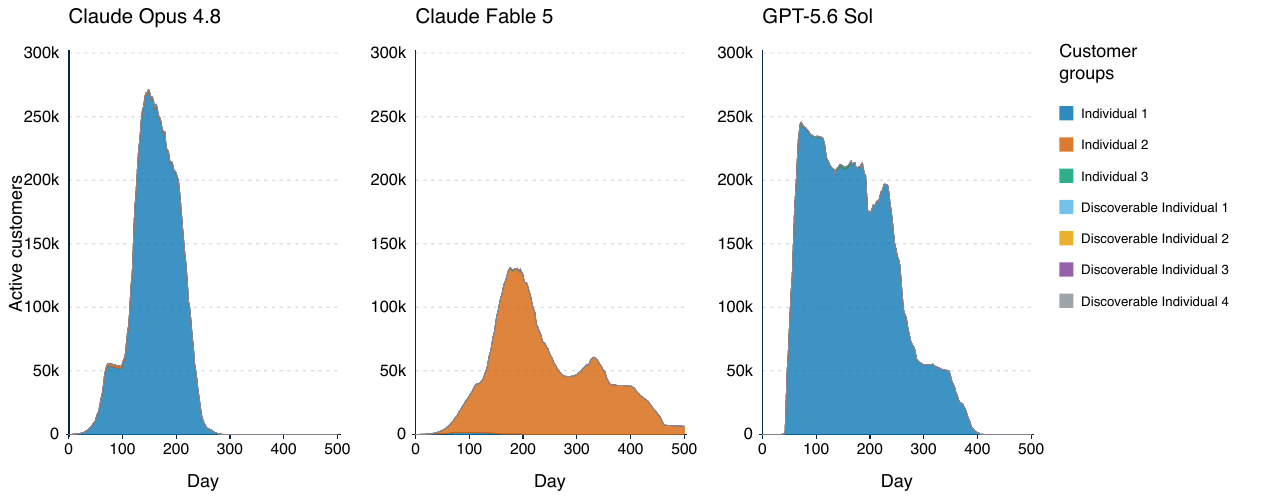}
  \caption{\textbf{Number of customers by customer group over time for the best runs of Claude Opus 4.8, Claude Fable 5, and GPT-5.6 Sol.} Claude Opus 4.8 obtains more customers initially and drops to zero customers mid-simulation, GPT-5.6 Sol scales Individual Group 1 fastest and gradually reaches zero customers by day 420, while Claude Fable 5 sustains a smaller customer base centered on Individual Group 2 through the end of the simulation. The three agents attain high final cash via distinct strategy styles. Discoverable customer groups are initially hidden to the agent and can only be discovered through paid market research.}
  \label{fig:customer-groups-by-model}
\end{figure}

\textbf{Agents attain high final cash through distinct strategies.} Claude Opus 4.8, Claude Fable 5, and GPT-5.6 Sol all finish above the starting cash in their best runs, but they attain this result through distinct strategies. In Fig.~\ref{fig:customer-groups-by-model}, we show their very different customer-base trajectories. Claude Opus 4.8 builds a larger early customer base and then drops to zero customers mid-simulation, GPT-5.6 Sol scales even earlier and gradually winds the customer base down to zero by day 420, while Claude Fable 5 sustains a smaller customer base centered on Individual Group 2 into the final week. Fig.~\ref{fig:memo-timeline} also shows that Claude Opus 4.8 pivots toward harvesting after its early growth phase, whereas Claude Fable 5 continues to revise its plan late in the run.

\textbf{Sophisticated analytics by top-performing agents.} In Fig.~\ref{fig:agent-code-example}, we show example code files that top-performing agents write and execute. Claude Opus 4.8 estimates which customers are likely to stay or leave by looking at who keeps paying and who cancels. Claude Fable 5 tracks when payments arrive, how costs accumulate, and forecasts how much cash remains in the final weeks under different customer-loss assumptions. These examples demonstrate initial signs of sophisticated planning and information acquisition.

\subsection{Measuring Drivers of Success and Failure}
\begin{figure}[!t]
  \centering
  \includegraphics[width=\textwidth]{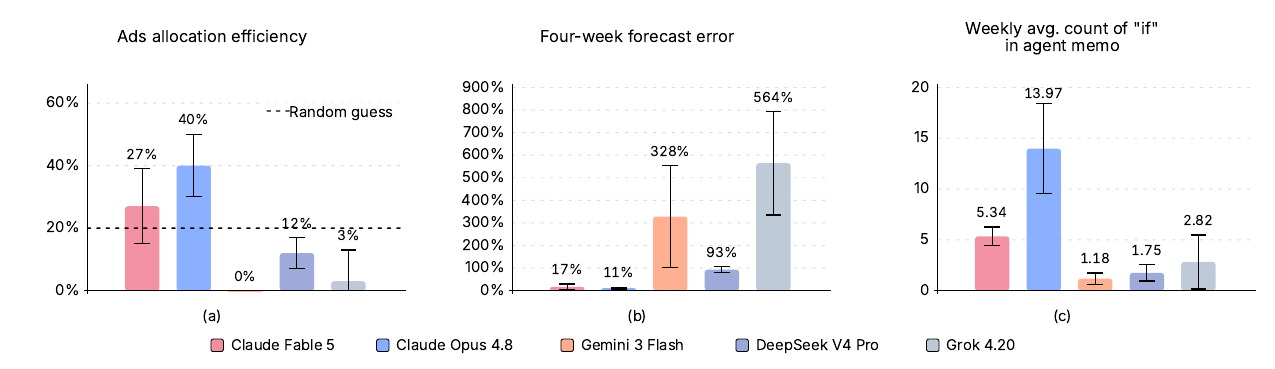}
  \caption{\textbf{Better-performing models excel along three behavioral measures}: (a) uncovering hidden ad-channel effectiveness and allocating spend to the best channel, (b) forecasting future cash, and (c) planning more extensively. We show mean and standard deviation of each measurement across runs in the plots above.}
  \label{fig:survival-vs-skill}
  \setcounter{subfigure}{0}
  \refstepcounter{subfigure}\label{fig:survival-vs-skill-a}
  \refstepcounter{subfigure}\label{fig:survival-vs-skill-b}
  \refstepcounter{subfigure}\label{fig:survival-vs-skill-c}
\end{figure}

\label{sec:failure-analysis}
While success in \textsc{CEO-Bench} requires multiple skills to work together, we conduct a preliminary analysis by isolating three behavioral measures and comparing them against agent performance. We compare quantitative measures for Claude Fable 5, Claude Opus 4.8, and three lower-performing comparison models in Fig.~\ref{fig:survival-vs-skill}, and explain each comparison further below.

\Needspace{0.2\textheight}
\begin{wrapfigure}{r}{0.50\textwidth}
  \centering
  \includegraphics[width=\linewidth]{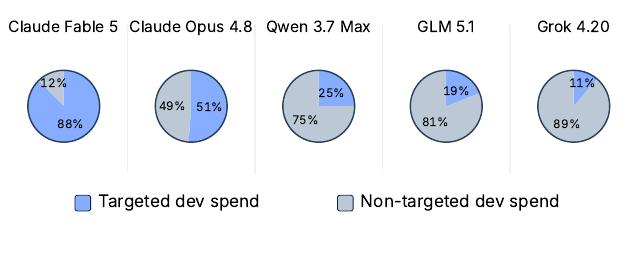}
  \caption{\textbf{Targeted development spending breakdown.} Better performing models tend to take advantage in actions targeted at specific customer groups more.}
  \label{fig:targeted-dev-spend}
\end{wrapfigure}

\textbf{Uncovering hidden information.} In \textsc{CEO-Bench}, each pair of ad channel and customer group has a different new-customer acquisition rate, emulating real-world heterogeneity. The acquisition rate is hidden from the agent, so the agent must uncover and use this information by analyzing customer acquisition history in databases. In Fig.~\ref{fig:survival-vs-skill-a}, we measure the average percentage of ad spending allocated to the best channel out of all ad spending for a customer group. Claude Opus 4.8 and Claude Fable 5 allocate more spend to the true best channel than the random-guessing baseline of $20\%$, while Gemini 3 Flash, DeepSeek V4 Pro, and Grok 4.20 fall below it.

\textbf{Seeing into the future.} In each simulated week, we ask the agent to submit a cash forecast four weeks into the future. Fig.~\ref{fig:survival-vs-skill-b} plots the percentage error between the submitted forecast and the realized cash balance four weeks later against the number of days before bankruptcy. We average data from the first four simulation weeks, when most models are still alive. Claude Opus 4.8 and Claude Fable 5 forecast future cash with much lower early error than Gemini 3 Flash, DeepSeek V4 Pro, and Grok 4.20, indicating a stronger understanding of world dynamics and delayed consequences.

\begin{figure}[t]
  \centering
  \includegraphics[width=\textwidth]{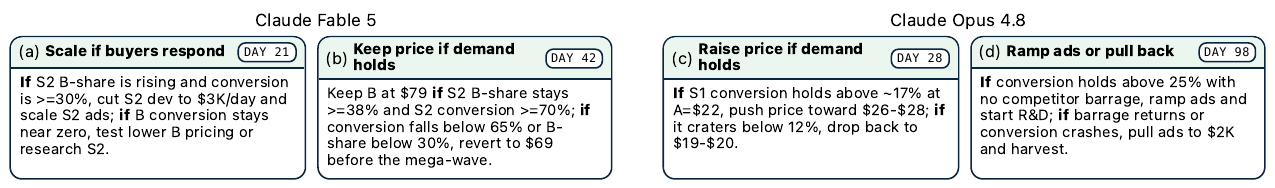}
  \caption{\textbf{Examples of planning in Claude Fable 5 and Claude Opus 4.8 memos.} The agents anticipate scenarios and solutions with ``if-then'' contingencies.}
  \label{fig:conditional-memo-examples}
\end{figure}

\textbf{Planning.} We find that the stronger runs frequently anticipate different future scenarios and build corresponding solutions in their memos, with examples shown in Fig.~\ref{fig:conditional-memo-examples}. The qualitative examples from Claude Fable 5 and Claude Opus 4.8 show explicit ``if-then'' contingencies, while the aggregate comparison in Fig.~\ref{fig:survival-vs-skill-c} shows that Claude Opus 4.8 and Claude Fable 5 use the word ``if'' more frequently than the lower-performing comparison models.

\textbf{Taking fine-grained actions.} Our simulator allows agents to take actions at a fine-grained level. For example, agents can decide customer-group specific product development strategy. Proper analysis of customer group information and targeted development can create advantages such as slower competitor catch-up. Fig.~\ref{fig:targeted-dev-spend} shows the dollar-weighted split between targeted and non-targeted development spending. Claude Fable 5 allocates 88\% of development dollars to targeted improvements, Claude Opus 4.8 allocates 51\%, Qwen 3.7 Max allocates 25\%, GLM 5.1 allocates 19\%, and Grok 4.20 allocates 11\%. This comparison demonstrates that stronger models acquire fine-grained understanding about world dynamics and make use of the understanding by taking granular actions more.
\section{Ablating Simulator and Agent Configurations}

We examine how simulation outcomes change when varying competitor and time-horizon configurations. We find that competitor difficulty provides an effective knob for tuning task difficulty, and our task remains challenging for existing models even over a short horizon. We also show that the choice of agent harness impacts results significantly.

\subsection{Ablating Competitor Difficulty}

\Needspace{0.16\textheight}

We ablate simulator difficulty by varying the competitor configuration while using Claude Sonnet 4.6 as the agent model. In \textsc{CEO-Bench}, the competitor raises customer expectations through both a preset stationary sequence and adaptive responses to agent actions. In the adaptive component, the competitor raises customer expectations by $u \cdot I$, where $u \sim U[0.2,0.5]$ and $I$ is the agent's cumulative quality improvement. We ablate simulator difficulty with the following settings: (1) stationary + adaptive competitor with $u \in \{0.1,0.2,0.3\}$; (2) stationary competitor only; (3) no competitor. The default line uses the best Claude Sonnet 4.6 trajectory under the standard configuration.

In Fig.~\ref{fig:ablation-simulator-configurations}(a), we show that reducing competitor strength generally makes the task easier, and removing the competitor makes the task significantly easier. The ablation shows that competitor strength can be an effective knob for tuning task difficulty, and the non-stationary environment is a crucial component of what makes the task challenging.
\subsection{Ablating Time-Horizon}

\begin{figure}[t]
  \centering
  \begin{subfigure}[t]{0.49\textwidth}
    \centering
    \includegraphics[width=\linewidth]{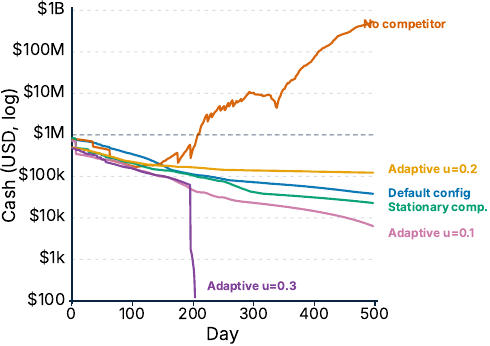}
  \end{subfigure}\hfill
  \begin{subfigure}[t]{0.49\textwidth}
    \centering
    \includegraphics[width=\linewidth]{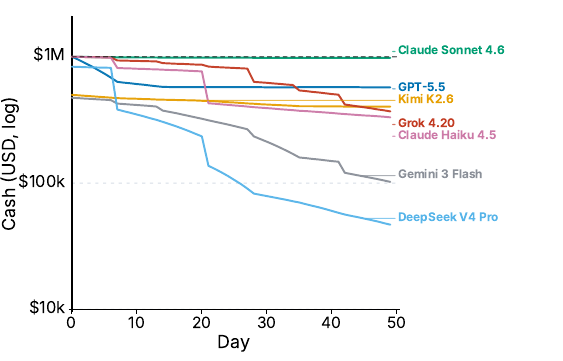}
  \end{subfigure}
  \caption{\textbf{Ablating simulator configurations.} (a) Weaker competitors generally make the task easier, and removing the competitor makes the task significantly easier. (b) Shortening the horizon to 50 days still leaves all evaluated models below their starting cash balance.}
  \label{fig:ablation-simulator-configurations}
\end{figure}

We examine whether agents behave differently when told to maximize cash balance over a shorter horizon. In this experiment, we change the simulation period to 50 days, one-tenth of the original simulation period. While the shortened horizon reduces challenges in long-term planning, Fig.~\ref{fig:ablation-simulator-configurations}(b) shows that no evaluated model finishes above its starting cash balance. This analysis reveals that most models today remain weak in orchestrating decisions toward a short-term goal.
\FloatBarrier
\Needspace{0.23\textheight}
\subsection{Ablating Agent Harness}

\begin{figure}[t]
  \centering
  \includegraphics[width=\textwidth]{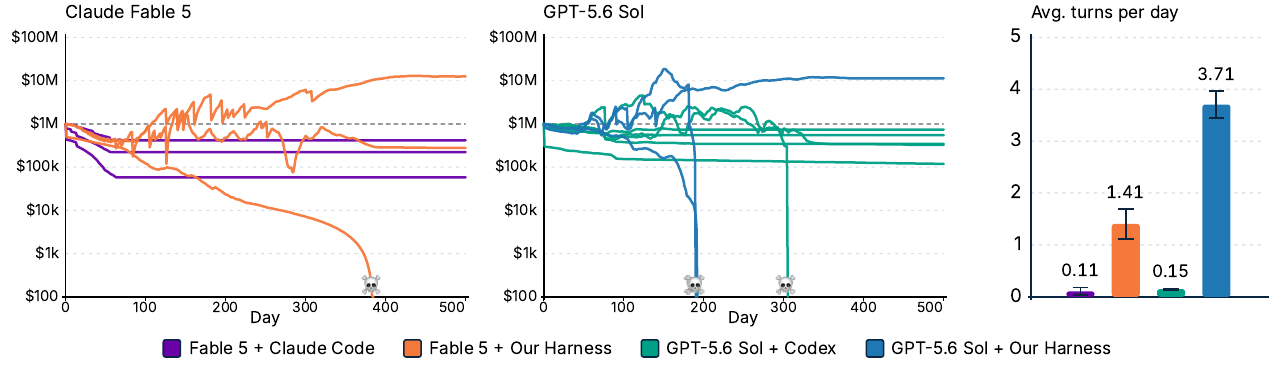}
  \caption{\textbf{Cash trajectories and turn frequency when ablating agent harnesses.} We compare Claude Fable 5 and GPT-5.6 Sol under our minimal terminal-using agent versus Claude Code and Codex respectively. Under Claude Code and Codex, the agents take fewer turns per simulated day and achieve inferior performance.}
  \label{fig:harness-comparison}
\end{figure}

\textsc{CEO-Bench} can easily evaluate any agent harness. While we obtain most results with a custom minimal terminal-using agent harness, we ablate popular agent harnesses while keeping the underlying model fixed. For Claude Fable 5, we compare results with Claude Code~\citep{anthropic2026claudecode}, and for GPT-5.6 Sol, we compare results with Codex~\citep{openai2026codexcli}. In Fig.~\ref{fig:harness-comparison}, we show that switching harnesses massively changes agent behaviors. Agents take significantly fewer turns per simulated day when using Claude Code and Codex, resulting in inferior performance. While we cannot access full implementation details of these harnesses, we hypothesize that the difference results from software engineering-oriented system prompts of these harnesses.

\section{Related Work}
\label{sec:related-work}

\paragraph{Language model evaluations.}
Language-model evaluation has moved from static knowledge and reasoning~\citep{hendrycks2021mmlu,srivastava2023bigbench,liang2023helm,rein2024gpqa} toward realistic agentic task execution~\citep{chen2021humaneval,jimenez2024swebench,zhou2024webarena,xie2024osworld,drouin2024workarena,trivedi2024appworld,yoran2024assistantbench,yao2024taubench,liu2024agentbench,ma2024agentboard}. Recent benchmarks have expanded evaluation scope to economically valuable deliverables in broad domains~\citep{patil2025bfcl,mialon2024gaia,chan2025mlebench,starace2025paperbench,miserendino2025swelancer,patwardhan2025gdpval}. However, their objectives usually terminate at a target state or one-shot deliverable.
\textsc{CEO-Bench} instead asks whether agents can sustain progress toward a distant objective as earlier decisions continue to shape later states.

% \paragraph{Memory and continual learning evaluations.}
% Memory and continual-learning benchmarks test whether models can retain, retrieve, update, or adapt from information accumulated over extended histories.
% LongBench, RULER, and LongMemEval evaluate long-context understanding or persistent chat memory~\citep{bai2023longbench,hsieh2024ruler,wu2025longmemeval}.
% Agentic-memory benchmarks move closer to interactive use, but still focus mainly on information storage and retrieval: MemoryAgentBench emphasizes storing and recalling evolving user information, while MemoryArena evaluates whether agents can use memory across multi-session tasks~\citep{hu2025memoryagentbench,he2026memoryarena}.
% Continual-learning and in-context reinforcement-learning benchmarks push further by testing whether models can convert past sequences or reward histories into productive signals for future behavior~\citep{laskin2023algorithm,wang2023trace,monea2024icrl}.
% These settings capture important ingredients of long-horizon agency, but they are typically narrow in goal and action space, and often lack lasting consequences across episodes.
% \textsc{CEO-Bench} instead asks whether accumulated evidence can be converted into operational decisions for steering a complex, changing system.
\paragraph{Long-horizon agent evaluation.}
Memory and continual-learning benchmarks test models' ability to retain information over time~\citep{bai2023longbench,hsieh2024ruler,wu2025longmemeval,hu2025memoryagentbench,he2026memoryarena,laskin2023algorithm,wang2023trace,monea2024icrl}, but they are often limited to information retrieval or a single static task. Long-horizon benchmarks extend evaluation from isolated tasks to processes that unfold over time~\citep{wu2023smartplay,xie2024travelplanner,xu2025theagentcompany,chan2025mlebench,starace2025paperbench,wang2025odysseybench,luo2025ultrahorizon,he2026ycbench}.
Most recently, Vending-Bench asks agents to run a vending machine over many days, and AccountingBench asks agents to close monthly books from real software company data~\citep{backlund2025vendingbench,backlund2025vendingbench2,penrose2025accountingbench}. However, they involve narrow operating problems, fewer coupled decisions, and largely stable or observable environments. \textsc{CEO-Bench} evaluates long-horizon agency in a broader operating setting where agents must coordinate pricing, growth, product, operations, communication, and enterprise sales under hidden state, noisy feedback, delayed consequences, and non-stationary market pressure in a consistent simulator. For example, we show in Appendix~\ref{app:vendingbench-comparison} that Vending-Bench allows models to accumulate successes relatively steadily, while our simulator requires an agent to make significant investments that only pay back much later, posing a stronger challenge to long-horizon planning. 
\section{Limitations and Conclusion}
\subsection{Limitations}
\label{sec:limitations}

We make our best effort to approximate real-world startup operations and challenges in \textsc{CEO-Bench}. However, discrepancies can still exist between reality and the approximation. For example, since we have not found a reliable way to evaluate a model's capability to propose qualitative changes to products, we simulate products using only a quality measure. In addition, to make each simulation run economically feasible, we limit the scope of possible actions and leave out aspects such as compliance, security, and fundraising.
\subsection{Conclusion}
\label{sec:conclusion}
    
\textsc{CEO-Bench} shows a gap between existing models' local tool competence and crucial sustained strategic skills: agents built on existing models can take plausible actions but fail when those actions must compound under delayed feedback, hidden state, and non-stationarity. To develop agents beyond isolated task executors, we need evaluations that ask whether they can organize evolving systems toward distant goals. \textsc{CEO-Bench} is one step toward that future: building agents and training models that do not merely answer requests, but help steer long-running organizations through uncertainty.

\section*{Acknowledgments}

We thank Modal for providing GPU resources for LLM inference. We thank Shuer Jiang, Boya Zeng, Sachin Konan, Taiming Lu, Linrong Cai, David Yin, Rahul Chalamala, Bryan Chiang, Luke Zeller, Yunyu Lin, Berkan Dokmeci, Bennett O'Brien, Ashank Tomar, and Ang Li for discussions and feedback. We thank Spencer Hong and The General Intelligence Company of New York for additional evaluations. KN acknowledges support from Schmidt Sciences.

%%%%%%%%%%%%%%%%%%%%%%%%%%%%%%%%%%%%%%%%%%%%%%%%%%%%%%%%%%%%

%%%%%%%%%%%%%%%%%%%%%%%%%%%%%%%%%%%%%%%%%%%%%%%%%%%%%%%%%%%%

\bibliographystyle{plainnat}
\bibliography{main}

%%%%%%%%%%%%%%%%%%%%%%%%%%%%%%%%%%%%%%%%%%%%%%%%%%%%%%%%%%%%

\newpage
\section*{\LARGE Appendix}

\vspace{.2cm}

\appendix

\section{Simulator Mechanics}
\label{app:simulator-details}
We describe full details of simulator mechanics in this section.

\subsection{Agent Commands and Observable State}
\label{app:python-interface-map}

\paragraph{Python action surface.}
The agent changes the company by importing functions from \verb|novamind_api| and executing Python. The active command groups are:
\begin{itemize}
    \item \textbf{\texttt{pricing}}: \verb|set_prices|, \verb|set_model_tiers|, \verb|set_usage_quotas|, and \verb|set_promotion|.
    \item \textbf{\texttt{marketing}}: \verb|set_daily_spend|, \verb|set_targeted_ad_spend|, and \verb|set_ads_strength|;\\
    \verb|set_lead_promotion| and \verb|post_social_media|.
    \item \textbf{\texttt{analytics}}: \verb|set_targeted_ops_spend| and \verb|set_targeted_dev_spend|.
    \item \textbf{\texttt{research}}: \verb|start_research_project| and \verb|list_research_projects|.
    \item \textbf{\texttt{market}}: \verb|research_market|, \verb|research_group|, \verb|get_market_overview|, and \verb|get_group_insights|.
    \item \textbf{\texttt{infrastructure}}: \verb|set_capacity_tier| and \verb|get_cost_info|.
    \item \textbf{\texttt{enterprise}}: \verb|send_enterprise_deal| and \verb|reject_enterprise_deal|.
    \item \textbf{Time and data access}: \verb|next_week| advances the simulator, \verb|query| reads the company database, and \verb|get_vars| reports runtime variables.
\end{itemize}
The interface is intentionally operational rather than omniscient: it exposes dashboards, database tables, social posts, and inbox messages, while leaving true preferences, satisfaction, competitor schedules, and hidden macro state latent.

\subsection{Customers, Plans, and Participation}
\label{app:customers-plans-participation}

\paragraph{Customer parameter sampling.}
Each customer belongs to a group $g$ and receives private parameters such as maximum willingness to pay, quality floor, quality ceiling, usage demand, ad sensitivity, support sensitivity, and enterprise negotiation traits. For a generic customer parameter $k$,
\begin{equation}
\theta_{i,k,t}=\operatorname{clip}\!\left(\tilde\theta_{i,k}+\Delta_{g,k,t}^{\mathrm{market}},\theta_{g,k}^{\min},\theta_{g,k}^{\max}\right),
\qquad
\tilde\theta_{i,k}\sim \mathcal D_{g,k}(\mu_{g,k},\sigma_{g,k}),
\end{equation}
where $\theta_{i,k,t}$ is customer $i$'s active value for parameter $k$, $\tilde\theta_{i,k}$ is the sampled base value, $\mathcal D_{g,k}$ is the configured group-level sampling distribution, $\mu_{g,k}$ and $\sigma_{g,k}$ are its location and spread parameters, $\Delta_{g,k,t}^{\mathrm{market}}$ is accumulated market drift, and $\theta_{g,k}^{\min},\theta_{g,k}^{\max}$ are clipping bounds. The distribution terms encode the fact that a market segment has a typical profile, while the sampled base value gives each customer an idiosyncratic budget, tolerance, or usage pattern. The drift term represents changing market conditions, such as customers becoming more demanding over time, and the clipping bounds keep preferences in plausible ranges. This mechanism simulates real customer cohorts: people in the same segment resemble one another, but they are not interchangeable, and their preferences can move as the market changes.

\paragraph{Customer participation curve.}
Each customer $i$ has maximum monthly willingness to pay $c_i$, minimum quality floor $q_i^{\min}$, quality ceiling $q_i^{\max}$, and low-price and high-price slopes $s_i^L,s_i^R$. For offered effective price $C$, define normalized price $x=C/c_i$, quality range $\Delta q_i=q_i^{\max}-q_i^{\min}$, and sigmoid $\sigma(z)=(1+e^{-z})^{-1}$. With configurable curve coefficients $\theta_Q^L,\theta_Q^R,\theta_Q^{\max},\omega_Q,k_Q,\psi_Q^{\min},\psi_Q^{\max}$,
\begin{equation}
\begin{aligned}
\psi_i(C)={}&
\omega_Q\,\sigma\!\left(k_Qs_i^L(x-\theta_Q^L)\right)
+(1-\omega_Q)\,\sigma\!\left(k_Qs_i^R(x-\theta_Q^R)\right),\\
Q_i^{\mathrm{req}}(C)={}&
q_i^{\min}+\Delta q_i\,\operatorname{clip}\!\left(\psi_i(C),\psi_Q^{\min},\psi_Q^{\max}\right).
\end{aligned}
\label{eq:customer-participation-curve}
\end{equation}
Here $\psi_i(C)$ is the customer's normalized required-quality score at price $C$; $x=C/c_i$ measures price relative to the customer's willingness to pay; $\Delta q_i$ is the customer's personal quality range; $\theta_Q^L$ and $\theta_Q^R$ locate the low- and high-price portions of the curve; $\omega_Q$ blends the two portions; $s_i^L$ and $s_i^R$ make some customers more price-sensitive than others; $k_Q$ controls global curvature; and $\psi_Q^{\min},\psi_Q^{\max}$ bound the score. The mechanic follows a participation-rule view of differentiated products: higher prices require higher perceived quality, and customers near their budget ceiling become more demanding~\citep{mussa1978monopoly}. A non-enterprise customer accepts plan $p$ only if $C_{i,p,t}\le \theta_Q^{\max}c_i$ and $Q_{i,p,t}^{\mathrm{perc}}\ge Q_i^{\mathrm{req}}(C_{i,p,t})$. This simulates the real purchasing rule that customers do not compare price in isolation; they ask whether the product feels good enough for what they are being charged.

\begin{figure}[h]
  \centering
  \includegraphics[width=0.72\textwidth]{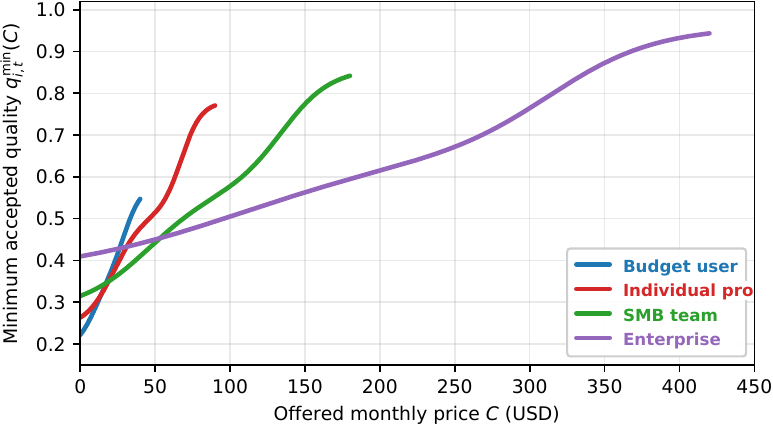}
  \caption{\textbf{Example customer participation curves.} Each curve represents the average participation behavior of a customer group. Each curve maps the offered monthly price $C$ to the minimum accepted quality $Q_i^{\mathrm{req}}(C)$ for a customer with different willingness to pay, quality floors, ceilings, and price-sensitivity slopes.}
  \label{fig:customer-participation-curves}
\end{figure}

\paragraph{Plan choice and billing-period switching.}
Let $\mathcal P$ be the active plan set. For customer $i$, the simulator evaluates all plans and chooses the acceptable plan with the largest surplus:
\begin{equation}
p_{i,t}^{*}=\arg\max_{p\in\mathcal P}
\left[Q_{i,p,t}^{\mathrm{perc}}-Q_i^{\mathrm{req}}(C_{i,p,t}^{\mathrm{eff}})\right],
\qquad
\mathcal A_{i,t}=\{p\in\mathcal P: C_{i,p,t}^{\mathrm{eff}}\le c_i,\; Q_{i,p,t}^{\mathrm{perc}}\ge Q_i^{\mathrm{req}}(C_{i,p,t}^{\mathrm{eff}})\}.
\end{equation}
The choice is valid only when $p_{i,t}^{*}\in\mathcal A_{i,t}$; if $\mathcal A_{i,t}$ is empty on a billing decision day, the customer cancels. $\mathcal P$ is the set of plans currently offered by the company, $C_{i,p,t}^{\mathrm{eff}}$ is the price after active promotions, $Q_{i,p,t}^{\mathrm{perc}}$ is perceived quality for plan $p$, $c_i$ is the customer's budget ceiling, and $\mathcal A_{i,t}$ is the acceptable-plan set. The surplus term measures how much perceived quality exceeds the customer's minimum requirement at that effective price. This creates a simple operator intuition: customers do not maximize quality alone or price alone; they choose the plan that clears their personal price-quality bar with the best headroom. In reality, this corresponds to monthly subscription review: customers may upgrade, downgrade, stay, or churn depending on the menu of plans available at renewal time.

\subsection{Product Quality, Usage, and Monetization}
\label{app:product-quality-usage-monetization}

\paragraph{Promotions and effective price.}
Promotions are additive across global, group, customer, and group-plan scopes. At billing time,
\begin{equation}
C_{i,p,t}^{\mathrm{eff}}=
\left[P_{p,t}
-\Pi_t^{\mathrm{global}}
-\Pi_{g,t}^{\mathrm{group}}
-\Pi_{i,t}^{\mathrm{customer}}
-\Pi_{g,p,t}^{\mathrm{group\text{-}plan}}
-\mathbb{1}\{\mathrm{first}_{i,t}\}\Pi_{g,t}^{\mathrm{lead}}\right]_+.
\end{equation}
Here $P_{p,t}$ is the listed price for plan $p$, $\Pi_t^{\mathrm{global}}$ is a site-wide discount, $\Pi_{g,t}^{\mathrm{group}}$ targets a customer segment, $\Pi_{i,t}^{\mathrm{customer}}$ targets an individual customer, $\Pi_{g,p,t}^{\mathrm{group\text{-}plan}}$ targets a segment-plan pair, $\Pi_{g,t}^{\mathrm{lead}}$ is a first-bill lead promotion, $\mathrm{first}_{i,t}$ marks the first billing event for a newly acquired customer, and $[\cdot]_+$ floors price at zero. Promotions therefore act as dollar discounts rather than hidden quality boosts; they can help acquisition or retention but directly reduce revenue. The mechanism simulates couponing, contract discounts, and introductory offers, where revenue changes immediately even though the product itself has not improved.

\paragraph{Usage and capacity.}
Each active subscriber has a daily usage draw $\tilde u_{i,t}$, plan quota $U_{p,t}$, billing-period cumulative usage $\bar U_{i,t}$, and weekly usage multiplier $W_t$. The realized daily usage is
\begin{equation}
u_{i,t}=\operatorname{round}\!\left(
\min\!\left(\tilde u_{i,t}W_t,\,[U_{p_i,t}-\bar U_{i,t}]_+\right)\right),
\qquad
U_t^{\mathrm{tot}}=\sum_i u_{i,t}.
\end{equation}
$u_{i,t}$ is the delivered usage units today, $\tilde u_{i,t}$ is the customer's latent demand, $W_t$ is a weekly demand multiplier, $U_{p_i,t}$ is the quota on the customer's active plan, $\bar U_{i,t}$ is already-consumed usage in the billing period, and $U_t^{\mathrm{tot}}$ is total platform load. The positive-part term enforces the remaining plan quota, and rounding maps continuous demand into discrete usage units. Plan quota also determines the quota fulfillment factor used in perceived quality below: lowering quota can reduce usage and compute cost, but when quota falls short of demand, it scales down the customer's entire experienced quality. The mechanic makes high-growth strategies stress infrastructure: more customers and higher quotas produce more usage, which can create overload if capacity is not upgraded. This simulates a real software business where usage is bursty, quotas cap consumption, and product-market success can turn into an infrastructure problem.

\paragraph{Service health.}
For capacity tier $\kappa_t$ with capacity $K_{\kappa_t}$ and operations spend $x_t^{\mathrm{ops}}$, the overload level and outage probability are
\begin{equation}
o_t=\left[\frac{U_t^{\mathrm{tot}}}{K_{\kappa_t}}-1\right]_+,
\qquad
P(\mathrm{outage}_t)=
\max\!\left(p_{\mathrm{out}}^{\min},p_{\mathrm{out}}^0\exp(-x_t^{\mathrm{ops}}/\chi_{\mathrm{ops}})\right)
\left(1+\beta_{\mathrm{out}}^o o_t\right).
\end{equation}
$o_t$ is zero when capacity covers demand and positive when load exceeds capacity; $K_{\kappa_t}$ is the capacity available under tier $\kappa_t$; $p_{\mathrm{out}}^0$ is baseline outage risk; $p_{\mathrm{out}}^{\min}$ is the reliability floor; $x_t^{\mathrm{ops}}$ is daily operations spend; $\chi_{\mathrm{ops}}$ controls diminishing returns to operations spend; and $\beta_{\mathrm{out}}^o$ makes overload increase outage risk. Operations spending therefore buys reliability, while capacity tier choices buy headroom. This simulates the operational reality that SRE effort and cloud capacity reduce incidents, but overloaded systems remain fragile and no team can drive outage risk exactly to zero.

\paragraph{Delivered quality.}
Delivered quality measures technical product quality before customer-specific perception effects:
\begin{equation}
Q_{g,p,t}^{\mathrm{del}} =
\left(q_0+b_t^{\mathrm{shared}}+b_{g,t}^{\mathrm{group}}\right)m_p
-\beta_o\,o_t-\beta_{\mathrm{out}}\,\mathbb{1}\{\mathrm{outage}_t\},
\end{equation}
where $q_0$ is baseline product quality, $b_t^{\mathrm{shared}}$ is shared quality from development and R\&D, $b_{g,t}^{\mathrm{group}}$ is targeted group quality for segment $g$, $m_p$ is the model-tier multiplier for plan $p$, $o_t$ is overload, and $\mathbb{1}\{\mathrm{outage}_t\}$ is the outage indicator. The coefficients $\beta_o$ and $\beta_{\mathrm{out}}$ translate infrastructure problems into quality loss. The product-quality terms model the underlying capability of the service, while the negative terms model degraded delivery. This separates product investment from delivery reliability: a strong product can still feel bad when overloaded, just as real customers judge both feature quality and whether the service actually works when they need it.

\paragraph{Development and R\&D.}
Daily development and targeted development add quality with diminishing returns:
\begin{equation}
\Delta b_t^{\mathrm{shared,dev}}=\beta_{\mathrm{dev}}\log(1+x_t^{\mathrm{dev}}/\chi_{\mathrm{dev}}),
\qquad
\Delta b_{g,t}^{\mathrm{target}}=\beta_{\mathrm{target}}\log(1+x_{g,t}^{\mathrm{target}}/\chi_{\mathrm{target}}).
\end{equation}
$x_t^{\mathrm{dev}}$ is global development spend, $x_{g,t}^{\mathrm{target}}$ is targeted development spend for group $g$, $\Delta b_t^{\mathrm{shared,dev}}$ is the daily shared-quality increment, $\Delta b_{g,t}^{\mathrm{target}}$ is the daily group-specific increment, $\beta_{\mathrm{dev}},\beta_{\mathrm{target}}$ convert spend into quality, and $\chi_{\mathrm{dev}},\chi_{\mathrm{target}}$ control diminishing returns. The logarithm makes the first dollars of engineering spend more productive than later dollars, reflecting coordination overhead and finite easy fixes. This simulates staffing and engineering allocation: basic improvements can be made quickly, but pushing quality further requires disproportionately more effort. R\&D projects are larger delayed improvements:
\begin{equation}
D_j^{\mathrm{R\&D}}\sim \mathcal D_{r_j}^{\mathrm{time}},
\qquad
G_j^{\mathrm{R\&D}}\sim \mathcal D_{r_j}^{\mathrm{quality}},
\qquad
b_{t+1}^{\mathrm{shared}}=b_t^{\mathrm{shared}}+\Delta b_t^{\mathrm{shared,dev}}+\sum_{j:\,t=t_j^{\mathrm{done}}}G_j^{\mathrm{R\&D}}+\epsilon_t^q.
\end{equation}
$j$ indexes a research project, $r_j$ is its tier, $D_j^{\mathrm{R\&D}}$ is completion delay, $\mathcal D_{r_j}^{\mathrm{time}}$ is the tier-specific time distribution, $G_j^{\mathrm{R\&D}}$ is its quality gain, $\mathcal D_{r_j}^{\mathrm{quality}}$ is the tier-specific gain distribution, $t_j^{\mathrm{done}}$ is the completion day, and $\epsilon_t^q$ is configured product-quality noise. The summation adds only projects that finish today, while daily development accumulates continuously. The design makes R\&D a delayed investment: it can move the global quality frontier, but it does not instantly solve today's churn risk. This simulates product roadmaps in which small engineering work compounds steadily, while larger research bets have uncertain delivery dates and payoffs.

\paragraph{In-app ads.}
In-app ad strength is additive across global, group, and customer settings and then log-scaled:
\begin{equation}
a_{i,t}^{\mathrm{eff}} =
\frac{
\log\!\left(\alpha_a+\kappa_a\,\operatorname{clip}(a_t^{\mathrm{global}}+a_{g,t}^{\mathrm{group}}+a_{i,t}^{\mathrm{customer}},a_{\min},a_{\max})\right)-\log \alpha_a
}{
\log\!\left(\alpha_a+\kappa_a a_{\max}\right)-\log \alpha_a
}.
\end{equation}
Here $a_t^{\mathrm{global}}$, $a_{g,t}^{\mathrm{group}}$, and $a_{i,t}^{\mathrm{customer}}$ are configured ad strengths at company, segment, and customer scope; $a_{\min},a_{\max}$ are bounds; $\alpha_a$ is a positive log offset; $\kappa_a$ controls how quickly raw ad strength saturates; and $a_{i,t}^{\mathrm{eff}}$ is the customer-visible ad load after clipping and saturation. The log scaling makes additional ad load less effective at the high end, matching the idea that an already ad-heavy product has limited extra monetization headroom. Ads create daily revenue
\begin{equation}
Y_{i,t}^{\mathrm{ads}}=\rho_i^{\mathrm{ads}}a_{i,t}^{\mathrm{eff}}n_i,
\end{equation}
where $\rho_i^{\mathrm{ads}}$ is customer $i$'s ad-revenue sensitivity and $n_i$ is seats. The same effective ad load subtracts from perceived quality. This creates a monetization tradeoff: ads are immediately lucrative but can reduce satisfaction and retention. The mechanism simulates ad-supported SaaS or freemium products, where more impressions produce revenue but also make the product feel noisier or less professional to some customers.

\paragraph{Perceived quality.}
Perceived quality is the utility-relevant quality experienced by customer $i$ after relationship, tenure, support, quota, and ad effects:
\begin{equation}
\begin{aligned}
Q_{i,t}^{\mathrm{perc}} ={}&
\underbrace{\phi_{i,p,t}^{\mathrm{quota}}}_{\text{quota factor}}
\Big[
Q_{g,p,t}^{\mathrm{del}}
+\beta_r(r_{i,t}-r_0)
+\beta_d\log\!\left(\alpha_d+d_{i,t}/d_0\right)
-\beta_I I_{i,t}\\
&-\eta_i^{\mathrm{ads}}a_{i,t}^{\mathrm{eff}}
\Big],
\end{aligned}
\end{equation}
where $r_{i,t}$ is relationship score, $r_0$ is neutral relationship, $d_{i,t}$ is days subscribed, $d_0$ is the tenure scale, $\alpha_d$ is the tenure log offset, $I_{i,t}$ is open-issue days, $U_{p,t}$ is plan quota, $u_i$ is sampled daily usage demand, $D_U$ converts daily demand to the quota period, and $a_{i,t}^{\mathrm{eff}}$ is effective ad load. Coefficients $\beta_r,\beta_d,\beta_I$ control relationship, tenure, and issue effects, while $\eta_i^{\mathrm{ads}}$ is the customer's ad-quality sensitivity. The quota fulfillment factor $\phi_{i,p,t}^{\mathrm{quota}}$ equals one when quota covers demand and falls toward zero when the plan delivers only a fraction of demand. Thus quota shortfalls scale the whole perceived-quality expression rather than subtracting a fixed penalty. This simulates the difference between engineering quality and customer experience: the same product can feel better to a long-tenured, well-supported customer and worse to a customer facing tickets, quotas, or intrusive ads.

\subsection{Satisfaction, Retention, and Support}
\label{app:satisfaction-retention-support}

\paragraph{Satisfaction.}
Instant satisfaction is quality surplus over the participation curve,
\begin{equation}
\tilde S_{i,t}=Q_{i,t}^{\mathrm{perc}}-Q_i^{\mathrm{req}}(C_{i,t}),
\end{equation}
and stored satisfaction is an exponential moving average with configurable inertia $\lambda_S$:
\begin{equation}
S_{i,t}=\lambda_SS_{i,t-1}+(1-\lambda_S)\tilde S_{i,t}.
\end{equation}
Here $\tilde S_{i,t}$ is today's surplus, $S_{i,t}$ is stored satisfaction, $C_{i,t}$ is the effective current price, $Q_i^{\mathrm{req}}(C_{i,t})$ is the quality the customer expects at that price, $Q_{i,t}^{\mathrm{perc}}$ is experienced quality, and $\lambda_S$ is satisfaction inertia. This is an expectancy-disconfirmation design: customers are satisfied when experience exceeds the paid-price expectation and dissatisfied when it falls short~\citep{oliver1980cognitive}. The moving average means customers remember recent experience instead of resetting each day, so a bad outage or a good support recovery can affect future behavior for multiple periods. Downstream rules weight negative satisfaction more strongly, consistent with loss aversion~\citep{kahneman1979prospect}. The mechanism simulates customer sentiment as a memory-bearing state rather than a one-day reaction.

\paragraph{Billing revenue and involuntary churn.}
On a billing day set by the billing period $D_{\mathrm{bill}}$, subscription revenue is
\begin{equation}
Y_t^{\mathrm{sub}}=\sum_{i\in\mathcal B_t} C_{i,p_i,t}^{\mathrm{eff}} n_i,
\end{equation}
where $\mathcal B_t$ is the set of subscribers billed on day $t$, $C_{i,p_i,t}^{\mathrm{eff}}$ is the effective price of customer $i$'s active plan, $p_i$ is the active plan, and $n_i$ is seats. Seat count multiplies revenue because an enterprise or team subscription pays for more users than an individual account. This simulates recurring subscription billing, where cash arrives in discrete renewal events rather than continuously every day. Before voluntary plan-choice churn, a group-level involuntary churn draw may occur:
\begin{equation}
\mu_{g,m}^{\mathrm{invol}}=\operatorname{clip}\!\left(\epsilon_{g,m}^{\mathrm{invol}},0,1\right),
\qquad
\epsilon_{g,m}^{\mathrm{invol}}\sim\mathcal N(\bar\mu_g^{\mathrm{invol}},\sigma_g^{\mathrm{invol}}),
\qquad
Z_{i,t}^{\mathrm{invol}}\sim\mathrm{Bernoulli}(\mu_{g,m}^{\mathrm{invol}}).
\end{equation}
$m$ indexes the billing period, $\epsilon_{g,m}^{\mathrm{invol}}$ is the period-specific involuntary churn rate before clipping, $\bar\mu_g^{\mathrm{invol}}$ and $\sigma_g^{\mathrm{invol}}$ are group-specific churn parameters, $\mu_{g,m}^{\mathrm{invol}}$ is the clipped probability used for group $g$ in period $m$, and $Z_{i,t}^{\mathrm{invol}}$ is the customer-level cancellation draw. This captures background churn such as procurement freezes, budget changes, company shutdowns, or stakeholder turnover that are not caused by the agent. Voluntary churn then follows the participation rule: if no plan clears the customer's curve, the customer cancels; if a different plan gives higher acceptable surplus, the customer switches. The mechanism simulates the fact that some churn is controllable through product and pricing, while some churn is exogenous noise in the customer base.

\paragraph{Support issue generation and resolution.}
For a subscriber with no open issue, the issue probability is
\begin{equation}
P(\mathrm{issue}_{i,t})=
\operatorname{clip}\!\left((p_0+p_S(S^{\mathrm{ref}}-S_{i,t})+p_{\mathrm{out}}\,\mathbb{1}\{\mathrm{outage}_t\})n_i,p_{\min},p_{\max}\right).
\end{equation}
$p_0$ is the base issue rate, $p_S$ converts low satisfaction into tickets, $S^{\mathrm{ref}}$ is the reference satisfaction level, $p_{\mathrm{out}}$ adds outage-driven issue risk, $\mathbb{1}\{\mathrm{outage}_t\}$ activates that risk on outage days, $n_i$ is seats, and $p_{\min},p_{\max}$ bound the probability. More seats create more chances for someone to hit a problem, while poor satisfaction and outages make support demand spike. This simulates customer-success queues where large accounts and unhappy users generate more tickets.

Open issues are resolved by operations pools. For pool $P$ with spend $x_P$, group $g$ members $n_{g,P}$, and pool size $|P|$,
\begin{equation}
N_{g,t}^{\mathrm{resolved}}\sim
\mathrm{Poisson}\!\left((b_P+\lambda_gx_P)\frac{n_{g,P}}{|P|}\right).
\end{equation}
$b_P$ is the pool's base resolution rate, $\lambda_g$ is group-specific operations efficiency, $x_P$ is spend assigned to support pool $P$, $n_{g,P}/|P|$ allocates capacity to group $g$ according to its share of the pool, and $N_{g,t}^{\mathrm{resolved}}$ is the number resolved. Global operations covers all open issues; targeted operations creates additional pools by group, plan, group-plan pair, or customer. Fast resolutions add relationship boosts, while unresolved issues increase open-issue days and decay relationship. The intuition is queue-based: more operations spend increases throughput, but only for customers covered by that pool. This simulates support staffing and escalation rules, where targeted customer-success effort can protect priority segments but cannot help customers outside the targeted pool.

\subsection{Reputation, Social Media, and Acquisition}
\label{app:reputation-social-acquisition}

\paragraph{Reputation impact.}
Each active customer contributes a daily reputation delta from satisfaction:
\begin{equation}
\delta_{i,t}^{\mathrm{rep}}=
\begin{cases}
\rho_+\,S_{i,t}, & S_{i,t}\ge S_0,\\
-\rho_-\,|S_{i,t}|, & S_{i,t}<S_0.
\end{cases}
\end{equation}
$S_0$ is neutral satisfaction, $\rho_+$ is the positive-reputation rate, $\rho_-$ is the negative-reputation rate, and $\delta_{i,t}^{\mathrm{rep}}$ is customer $i$'s daily reputation contribution. Negative satisfaction is allowed to have a different slope from positive satisfaction so that bad experiences can be more reputationally damaging than good experiences are helpful. This simulates word-of-mouth asymmetry in real markets, where angry customers often spread more salient feedback than mildly satisfied customers. For group $g$,
\begin{equation}
\Delta R_{g,t}=\frac{\sum_{i\in g}\delta_{i,t}^{\mathrm{rep}}}{\max(N_g,N_{\min})}\log_{\nu_N}(\max(N_g,N_{\min})),
\qquad
R_{g,t+1}=\operatorname{clip}(R_{g,t}+\Delta R_{g,t},R_{\min},R_{\max}).
\end{equation}
$N_g$ is the active subscriber count, $N_{\min}$ is the small-sample normalizer, $\nu_N$ controls logarithmic scale, $R_{g,t}$ is group reputation, and $R_{\min},R_{\max}$ bound reputation. The averaging term prevents a single customer from dominating a large segment, while the logarithmic factor lets larger customer bases produce more visible aggregate reputation movement. This simulates customer reviews and public sentiment accumulating within a market segment. Cancellations add event damage
\begin{equation}
D_{i,t}^{\mathrm{cancel}}=\eta_D(\beta_D+\xi_t)\left(\alpha_D+\chi_D\min(S_{i,t},S_0)^2\right)
\frac{\log_{\nu_N}(\max(N_g,N_{\min}))}{\max(N_g,N_{\min})},
\qquad
\xi_t\sim\mathrm{Uniform}(\xi_{\min},\xi_{\max}),
\end{equation}
where $\eta_D,\beta_D,\alpha_D,\chi_D$ are configurable damage coefficients, $\min(S_{i,t},S_0)^2$ makes very negative satisfaction especially costly, $\xi_t$ is event noise, and $D_{i,t}^{\mathrm{cancel}}$ is the reputation hit from cancellation. This makes visible churn more damaging when the customer was very unhappy. The mechanism simulates public cancellations, angry posts, and negative references that can hurt a brand beyond the lost subscription revenue.

\paragraph{Cross-group reputation spillovers.}
Discovered groups receive spillovers from related groups:
\begin{equation}
R_{h,t+1}\leftarrow
\operatorname{clip}\!\left(R_{h,t+1}+\zeta_R W_{g,h}\Delta R_{g,t},R_{\min},R_{\max}\right).
\end{equation}
$W_{g,h}$ is the influence from group $g$ to group $h$, $\zeta_R$ scales spillover, $\Delta R_{g,t}$ is the reputation change in the source group, and the clipping bounds keep the recipient group's reputation in range. The mechanic makes reputation networked: enterprise failures can affect nearby enterprise groups or adjacent market segments. This simulates reference networks, professional communities, and social adjacency, where the experience of one group can change expectations in a related group even before those customers use the product.

\paragraph{Customer and agent social media.}
Customer social-media candidates are weighted by satisfaction extremity, negative satisfaction, recent satisfaction change, active service events, influencer status, and seat count:
\begin{equation}
w_{i,t}^{\mathrm{post}}=
n_i\,\omega_g^{\mathrm{inf}}\omega_{i,t}^{\mathrm{new}}
\cdot\left(1+\alpha_S|S_{i,t}|\right)
\cdot\left(1+\alpha_-[-S_{i,t}]_+^2\right)
\cdot\left(1+\alpha_{\Delta}|\Delta S_{i,t}|\right)
\cdot\left(1+\alpha_E|\mathcal E_{i,t}|\right).
\end{equation}
$w_{i,t}^{\mathrm{post}}$ is the post-sampling weight, $n_i$ is seats, $\omega_g^{\mathrm{inf}}$ is the group influence multiplier, $\omega_{i,t}^{\mathrm{new}}$ is the new-customer multiplier, $\alpha_S$ weights satisfaction extremity, $\alpha_-$ gives extra weight to negative satisfaction, $\alpha_{\Delta}$ weights recent satisfaction changes, $\alpha_E$ weights active service events, $\Delta S_{i,t}$ is the satisfaction change, and $\mathcal E_{i,t}$ is the set of active quality events such as outage, overload, issue, or quota frustration. The simulator samples up to $K_{\mathrm{post}}$ posts per day from these candidates, so social media is a noisy but informative public signal rather than a complete survey. This simulates the selection bias of public feedback: large, influential, newly acquired, or upset customers are more likely to be heard than a random satisfied user.

Agent-authored posts are judged per discovered group. For group $g$, let $e_{g,t}^{\mathrm{agent}}\in[-1,1]$ be the judged reaction score. The social multiplier entering acquisition is
\begin{equation}
A_{g,t}=\operatorname{clip}\!\left(A_0+\alpha_A e_{g,t}^{\mathrm{agent}}V_{g,t},A_{\min},A_{\max}\right),
\end{equation}
where $e_{g,t}^{\mathrm{agent}}$ is the judged reaction of group $g$ to the agent's post, $V_{g,t}$ is exposure, $A_0$ is neutral social effect, $\alpha_A$ converts reaction-weighted exposure into lead impact, and $A_{\min},A_{\max}$ bound the multiplier. Public communication can therefore help or hurt growth depending on how each group reacts. This simulates product marketing and public relations: a message that resonates with one segment can accelerate acquisition, while a poorly received message can suppress demand.

\paragraph{Daily new-customer generation.}
For discovered target group $g$, expected leads are
\begin{equation}
\lambda_{g,t}=
R_{g,t}D_{g,t}C_tM_{g,t}A_{g,t}Z_t
\left(
\sum_c \frac{x_{c,g,t}L_{c,g,t}}{x_{\mathrm{ad}}}
+\sum_h N_{h,t}W^{\mathrm{net}}_{h,g}
\right),
\end{equation}
where $R_{g,t}$ is reputation, $D_{g,t}$ is market availability, $C_t$ is the calendar-cycle multiplier, $M_{g,t}$ is the macro lead multiplier, $A_{g,t}$ is the agent-social multiplier, $Z_t$ is the active demand-surge multiplier, $x_{c,g,t}$ is channel spend, $L_{c,g,t}$ is leads per reference ad spend $x_{\mathrm{ad}}$, $N_{h,t}$ is subscribers in group $h$, and $W^{\mathrm{net}}_{h,g}$ is the referral matrix. The calendar-cycle multiplier makes demand oscillate through recurring seasonal cycles, so otherwise identical ad spend can perform better or worse depending on timing. The macro multiplier captures broad economic expansion or contraction; the social and surge multipliers capture communication effects and temporary external demand spikes; and the referral term captures word of mouth from existing subscribers. Market saturation is
\begin{equation}
D_{g,t}=\left[D_0-\left(\frac{N_{g,t}}{\mathrm{cap}_{g,0}(\alpha_{\mathrm{cap}}+\gamma_gt/Y_{\mathrm{cap}})}\right)^{\nu_D}\right]_+,
\qquad
n_{g,t}\sim\mathrm{Poisson}([\lambda_{g,t}]_+).
\end{equation}
$D_0$ is baseline availability, $N_{g,t}$ is the current number of customers in group $g$, $\mathrm{cap}_{g,0}$ is initial market capacity, $\alpha_{\mathrm{cap}}$ is the baseline capacity scale, $\gamma_g$ is group capacity growth, $Y_{\mathrm{cap}}$ is the time scale for capacity growth, $\nu_D$ controls saturation curvature, $\lambda_{g,t}$ is expected leads, and $n_{g,t}$ is realized leads. The positive-part operator prevents negative availability, and the Poisson draw converts the expected funnel volume into noisy realized leads. This combines paid acquisition, word of mouth, reputation, macro conditions, seasonal demand, temporary shocks, and finite market size. The mechanism simulates a real go-to-market funnel where the same budget can yield different outcomes depending on brand, timing, segment saturation, and randomness.

\paragraph{Demand surges.}
External demand surges are temporary acquisition shocks. For each active surge $s$,
\begin{equation}
Z_t=\prod_{s\in\mathcal S_t} z_s,
\qquad
z_s\sim\mathcal D_s^{\mathrm{surge}},
\qquad
t\in[t_s^{\mathrm{start}},t_s^{\mathrm{end}}).
\end{equation}
$\mathcal S_t$ is the set of active surges, $z_s$ is the surge lead multiplier sampled from surge distribution $\mathcal D_s^{\mathrm{surge}}$, and $t_s^{\mathrm{start}},t_s^{\mathrm{end}}$ are its active days. The product over active surges allows multiple external events to stack. Surges create temporary windows where growth is easier, but the agent must still have pricing, quality, and capacity to retain the acquired customers. This simulates events such as press attention, industry shifts, or sudden demand spikes that increase inbound interest without guaranteeing durable revenue.

\subsection{Market Discovery and Non-Stationarity}
\label{app:market-discovery-nonstationarity}

\paragraph{Market and group research.}
Market research reveals new customer groups, while group research increases the information level for a known group after a delay:
\begin{equation}
\ell_{g,t+D_{g,\ell}^{\mathrm{research}}}^{\mathrm{info}}
=\max(\ell_{g,t}^{\mathrm{info}},\ell^{\mathrm{target}}),
\qquad
D_{g,\ell}^{\mathrm{research}}\sim\mathcal D_{g,\ell}^{\mathrm{research}}.
\end{equation}
$\ell_{g,t}^{\mathrm{info}}$ is the agent-visible information level for group $g$, $\ell^{\mathrm{target}}$ is the requested level, $D_{g,\ell}^{\mathrm{research}}$ is the configured completion delay, and $\mathcal D_{g,\ell}^{\mathrm{research}}$ is the delay distribution for that group and research depth. The max operator means research can raise the information level but cannot erase already acquired knowledge. Results snapshot current market conditions when the research completes. The purpose is to make information acquisition an operational choice with time cost rather than a free static table. This simulates customer discovery, analyst work, and market research projects that improve visibility only after a delay and may already be slightly stale when delivered.

\paragraph{Competitor events.}
Competitor events are disabled before $t_{\mathrm{comp}}^{\mathrm{start}}$ and after $t_{\mathrm{comp}}^{\mathrm{end}}$. Let $\tau$ be the last event day. The mean interval is
\begin{equation}
\bar\Delta_t =
\begin{cases}
m_\Delta\Delta, & t<t_\Delta^{\mathrm{switch}},\\
\Delta, & t\ge t_\Delta^{\mathrm{switch}},
\end{cases}
\end{equation}
with a separately configured minimum interval $\Delta_{\min}$. If $t-\tau<\Delta_{\min}$, no event occurs; otherwise
\begin{equation}
E_t\sim \mathrm{Bernoulli}(1/\bar\Delta_t).
\end{equation}
When an event occurs, a base boost is sampled and scaled over the run:
\begin{equation}
B_t^{\mathrm{sample}} =
\operatorname{clip}\!\left(\mathrm{LogNormal}(\mu_B,\sigma_B),B_{\min},B_{\max}\right)
\left(\alpha_B+\rho_B\frac{t-t_B^{\mathrm{start}}}{T_B^{\mathrm{ramp}}}\right).
\end{equation}
$t_{\mathrm{comp}}^{\mathrm{start}}$ and $t_{\mathrm{comp}}^{\mathrm{end}}$ bound the active competitor window, $\tau$ is the last event day, $\bar\Delta_t$ is the current mean interval between competitor events, $E_t$ is the event indicator, $m_\Delta$ slows early events, $\Delta$ is the baseline interval, $t_\Delta^{\mathrm{switch}}$ is the interval switch day, and $\Delta_{\min}$ prevents events from arriving too close together. $\mu_B,\sigma_B$ parameterize event size, $B_{\min},B_{\max}$ bound it, $t_B^{\mathrm{start}}$ is the boost-ramp start day, and $\alpha_B,\rho_B,T_B^{\mathrm{ramp}}$ control magnitude scaling over the run. Together these terms simulate rival launches that are not perfectly periodic, but become more serious as the market matures. The simulator also models adaptive competitor catch-up to the agent's unreleased global development and R\&D gains. Let $H_t^{\mathrm{global}}$ be the unreleased shared-quality bank. With $u_t^{\mathrm{global}}\sim\mathrm{Uniform}(u_{\min}^{\mathrm{global}},u_{\max}^{\mathrm{global}})$,
\begin{equation}
B_t=\max\!\left(B_t^{\mathrm{sample}},u_t^{\mathrm{global}}H_t^{\mathrm{global}}\right),
\qquad
H_{t+}^{\mathrm{global}}=
H_t^{\mathrm{global}}-\mathbb{1}\{u_t^{\mathrm{global}}H_t^{\mathrm{global}}>B_t^{\mathrm{sample}}\}\,u_t^{\mathrm{global}}H_t^{\mathrm{global}} .
\end{equation}
Thus $B_t$ is the applied competitor quality boost, $B_t^{\mathrm{sample}}$ is the exogenous shock, $u_t^{\mathrm{global}}$ is the random catch-up fraction, and $H_{t+}^{\mathrm{global}}$ is the remaining shared-quality bank after the event. The event boost is at least the stationary sampled shock but can be larger when the agent has accumulated large unreleased global improvements. If the adaptive term wins, the competitor consumes that fraction of the bank. This simulates competitors copying or matching broadly visible product improvements: large general advances can invite stronger competitive responses than small incremental changes. Targeted development has a parallel group-specific adaptive component:
\begin{equation}
D_{g,t}^{\mathrm{target}}=v_{g,t}H_{g,t}^{\mathrm{target}},
\qquad
v_{g,t}\sim\mathrm{Uniform}(v_{\min},v_{\max}),
\qquad
H_{g,t+}^{\mathrm{target}}=H_{g,t}^{\mathrm{target}}-D_{g,t}^{\mathrm{target}},
\end{equation}
where $H_{g,t}^{\mathrm{target}}$ is the unreleased targeted-development bank, $v_{g,t}$ is the group catch-up fraction sampled between $v_{\min}$ and $v_{\max}$, $D_{g,t}^{\mathrm{target}}$ is the targeted drift shock, and $H_{g,t+}^{\mathrm{target}}$ is the remaining targeted bank after catch-up. The applied event shifts all customers' curves upward by adding $B_t$ to global quality expectation, $\kappa_gB_t$ to group $g$'s expectation, and $D_{g,t}^{\mathrm{target}}$ to group $g$'s expectation. Here $\kappa_g$ is group competitor reactivity. The intuition is that competitors are partly exogenous and partly responsive: broad R\&D attracts broad catch-up, while targeted work is harder for competitors to fully copy but can still leak into group expectations. This simulates real competitive pressure where segment-specific improvements can create more durable advantage than broad, easily observed product gains.

\paragraph{Macroeconomic schedule.}
The hidden macro state is an Ornstein--Uhlenbeck process around a sinusoidal PMI cycle~\citep{ornstein1930brownian,szimayer2004testing,koenig2002pmi}:
\begin{equation}
\begin{aligned}
\mu_t={}&\mu_{\mathrm{PMI}}+A_{\mathrm{PMI}}\sin\!\left(2\pi t/P_{\mathrm{PMI}}+\phi\right),\\
\mathrm{PMI}_{t+1}={}&
\operatorname{clip}\!\left(
\mathrm{PMI}_t+\eta_{\mathrm{PMI}}(\mu_t-\mathrm{PMI}_t)+\sigma_{\mathrm{PMI}}\epsilon_t,
\mathrm{PMI}_{\min},\mathrm{PMI}_{\max}\right).
\end{aligned}
\end{equation}
with $\epsilon_t\sim\mathcal N(\mu_\epsilon,\sigma_\epsilon^2)$. $\mu_t$ is the current cycle target, $\mu_{\mathrm{PMI}}$ is the baseline PMI, $A_{\mathrm{PMI}}$ is cycle amplitude, $P_{\mathrm{PMI}}$ is cycle period, $\phi$ is phase, $\eta_{\mathrm{PMI}}$ is mean-reversion speed, $\sigma_{\mathrm{PMI}}$ is shock scale, and $\mathrm{PMI}_{\min},\mathrm{PMI}_{\max}$ bound the index. Intuitively, demand oscillates through a PMI-like expansion-and-contraction cycle, while the OU update makes temporary surprises fade back toward the current cycle phase. This simulates macroeconomic background conditions that drift gradually but also contain noise. For each customer group and macro-sensitive dimension $d$,
\begin{equation}
M_{g,d,t}=\max\!\left(M_{\min},M_0+\beta_{g,d}\frac{\mathrm{PMI}_t-\mathrm{PMI}_0}{\mathrm{PMI}_0}\right).
\end{equation}
$M_{g,d,t}$ is the macro multiplier, $M_0$ is neutral, $M_{\min}$ is its floor, $\beta_{g,d}$ is group sensitivity, and $\mathrm{PMI}_0$ is the neutral PMI reference. The dimension $d$ can represent macro-sensitive quantities such as lead flow, willingness to pay, or enterprise deal velocity, and different groups react with different sensitivities. The simulator uses real-time hidden PMI internally, while the agent sees only period averages published every $D_{\mathrm{publish}}$ days with delay $D_{\mathrm{delay}}$. This simulates management under delayed economic indicators: the world has already moved before the agent receives clean macro data.

\subsection{Enterprise Sales and Negotiation}
\label{app:enterprise-sales-negotiation}

\paragraph{Offer evaluation.}
Enterprise customers evaluate up to $K_{\mathrm{offer}}$ offered $(\mathrm{plan},\mathrm{price})$ options and choose the one with highest
\begin{equation}
S_i^{\mathrm{offer}}=Q_{i,p,t}^{\mathrm{perc}}-Q_i^{\mathrm{req}}(C),
\end{equation}
where $K_{\mathrm{offer}}$ is the maximum number of options the agent can present, $Q_{i,p,t}^{\mathrm{perc}}$ is the perceived quality of the offered plan, $Q_i^{\mathrm{req}}(C)$ is the required quality at price $C$, and $S_i^{\mathrm{offer}}$ is offer surplus. Positive surplus means the proposed plan clears the customer's price-quality bar. If the best offer is positive, the customer accepts. This simulates enterprise procurement evaluating a small menu of contract options rather than passively accepting a list price. Otherwise, the simulator computes
\begin{equation}
C_i^{\max}(Q)=\max\{C\le c_i: Q_i^{\mathrm{req}}(C)\le Q\}
\end{equation}
where $C_i^{\max}(Q)$ is the highest price the customer would accept at perceived quality $Q$, $c_i$ is the customer's budget ceiling, and the max operator searches for the price that is still justified by the offered quality. This is the reservation price implied by the same participation curve used for self-serve customers. This simulates a procurement ceiling: the customer may negotiate, but there is a maximum contract value that the perceived product quality can support. With sampled initial factor $f_i$ and configured counter-offer decay $\gamma_\alpha$,
\begin{equation}
C_{i,r}^{\mathrm{counter}}
=C_i^{\max}-\gamma_\alpha^r\left(C_i^{\max}-f_iC_i^{\max}\right),
\end{equation}
where $C_{i,r}^{\mathrm{counter}}$ is the counter-offer on turn $r$, $f_i$ is the initial counter-offer fraction of the customer's maximum acceptable price, and $\gamma_\alpha$ controls how quickly later counter-offers approach that maximum. Early counter-offers start below true willingness to pay; later turns move toward the customer's maximum acceptable price. If the configured maximum turn count is exceeded, the customer stops responding. This simulates negotiation anchoring: buyers may reveal willingness to pay gradually rather than immediately offering their ceiling. Reply delays are also stochastic:
\begin{equation}
d_{i,r}^{\mathrm{reply}}\sim\mathcal D_i^{\mathrm{reply}}\!\left(\bar d_i/M_{g,\mathrm{deal},t},\sigma_i^d\right),
\end{equation}
where $d_{i,r}^{\mathrm{reply}}$ is the delay before the customer responds, $\bar d_i$ and $\sigma_i^d$ are customer response-time parameters, $\mathcal D_i^{\mathrm{reply}}$ is the customer-specific reply-delay distribution, and $M_{g,\mathrm{deal},t}$ is the macro deal-velocity multiplier. Stronger macro conditions shorten expected delay by increasing the denominator, while slow conditions stretch sales cycles. This makes enterprise sales slower and less certain than self-serve conversion while still following the same underlying price-quality logic. The mechanism simulates procurement latency, stakeholder review, and macro-sensitive sales velocity.

\subsection{Costs and Cash Flow}
\label{app:costs-cash-flow}

\paragraph{Daily costs.}
Daily operating cost combines fixed infrastructure, variable compute, operations, development, advertising, targeted actions, lead acquisition, and research charges:
\begin{equation}
\begin{aligned}
K_t^{\mathrm{cost}}=
{}&K_{\kappa_t}^{\mathrm{capacity}}
+\sum_p \chi_p^{\mathrm{usage}}U_{p,t}^{\mathrm{use}}
+x_t^{\mathrm{ops}}+x_t^{\mathrm{dev}}
+X_t^{\mathrm{target\text{-}ops}}
+\sum_{g}x_{g,t}^{\mathrm{target\text{-}dev}}
+\sum_{c,g}x_{c,g,t}^{\mathrm{ads}}
\\
&+N_t^{\mathrm{lead}}c^{\mathrm{lead}}
+K_t^{\mathrm{market}}+K_t^{\mathrm{group}}+K_t^{\mathrm{project}}
\end{aligned}
\end{equation}
$K_{\kappa_t}^{\mathrm{capacity}}$ is the fixed cost of capacity tier $\kappa_t$, $\chi_p^{\mathrm{usage}}$ is per-usage cost for plan $p$, $U_{p,t}^{\mathrm{use}}$ is usage on that plan, $x_t^{\mathrm{ops}}$ and $x_t^{\mathrm{dev}}$ are global operations and development spend, $X_t^{\mathrm{target\text{-}ops}}$ is targeted support spend, $\sum_gx_{g,t}^{\mathrm{target\text{-}dev}}$ is targeted development spend across groups, $\sum_{c,g}x_{c,g,t}^{\mathrm{ads}}$ is acquisition advertising spend across channels and groups, $N_t^{\mathrm{lead}}c^{\mathrm{lead}}$ is the per-lead acquisition charge, and the three $K$ terms are market research, group research, and research-project charges paid that day. This cost equation simulates the operating budget of a software company, where fixed infrastructure, variable compute, staffing, marketing, and research all consume cash through different channels. The cash update is
\begin{equation}
B_{t+1}=B_t+Y_t^{\mathrm{sub}}+\sum_iY_{i,t}^{\mathrm{ads}}-K_t^{\mathrm{cost}},
\end{equation}
where $B_t$ is company cash, $Y_t^{\mathrm{sub}}$ is subscription revenue, $\sum_iY_{i,t}^{\mathrm{ads}}$ is in-product ad revenue, and $K_t^{\mathrm{cost}}$ is total operating cost. This is the main strategic coupling in the simulator: growth, quality, and reliability can create future revenue, but they consume cash immediately. The mechanism simulates startup runway management, where the agent must decide when to burn cash for future growth and when to preserve liquidity to avoid bankruptcy.

\section{Rule-Based Baseline Strategy and Configuration Search}
\label{app:rule-based-baseline}

We use the rule-based baseline as a non-LLM point of comparison for the agent results in Section~\ref{sec:results-overview}. The baseline is intentionally simple: it commits to a fixed pricing book, fixed product-quality and advertising spend levels, and a fixed customer targeting rule. It does not use market research, enterprise negotiation, social media analysis, promotions, or language-model calls. 

At the start of a run, the policy sets prices, model tiers, and usage quotas from one price book. During each simulated week, it selects target customer groups from its target rule. If cash is above a configured floor, it applies a small global development spend of \$200/day, a fixed targeted development spend for each selected group, and, starting on day 20, a fixed targeted advertising spend for each selected group. Advertising is placed on the highest-yield channel for the selected group under the simulator's fixed channel table. Operations spend is set to $\max(100,0.05n_t)$ dollars/day, where $n_t$ is the current active subscriber count. The policy adjusts capacity by at most one tier per week toward the cheapest tier whose capacity covers recent average usage at 80\% utilization. This simulates a simple operating playbook: spend proportionally to current scale, focus on a small target market, and avoid complex adaptation or hidden-state inference.

The configuration search is the Cartesian product of the options in Table~\ref{tab:rule-baseline-search-space}, giving 24 configurations. Each configuration is evaluated for the full 500-day simulation with seed 42. The best configuration uses the mid price book, targets S1 only, uses the heavy spend package, and has a \$100K cash floor; the \$300K cash floor gives the same result for this seed. The traced replay of this configuration ends with \$15.76M in cash.

\begin{table}[h]
  \centering
  \tablefontsize
  \begin{tabular}{p{0.23\linewidth}p{0.70\linewidth}}
    \toprule
    Search dimension & Options \\
    \midrule
    Price book &
    \textbf{cheap}: A=\$8/T1/50K tokens, B=\$18/T2/200K tokens, C=\$40/T3/1M tokens;
    \textbf{mid}: A=\$12/T1/60K tokens, B=\$25/T2/250K tokens, C=\$55/T4/1.2M tokens. \\
    Target rule &
    \textbf{S1 only}: target S1 throughout the run;
    \textbf{S1 then S3}: target S1 initially, then target both S1 and S3 from day 30 onward. \\
    Spend package &
    \textbf{light}: S1 dev/ad=\$2K/\$250 per day, S3 dev/ad=\$1.5K/\$600 per day;
    \textbf{medium}: S1 dev/ad=\$4K/\$500, S3 dev/ad=\$3K/\$1.2K;
    \textbf{heavy}: S1 dev/ad=\$6K/\$750, S3 dev/ad=\$4.5K/\$1.8K. \\
    Cash floor &
    Stop optional global development, targeted development, and advertising when cash is at or below either \$100K or \$300K. \\
    Fixed choices &
    Advertising begins on day 20; capacity targets 80\% utilization; market discovery and enterprise actions are disabled; default competitor events remain active. \\
    \bottomrule
  \end{tabular}
  \caption{\textbf{Search space for the rule-based baseline.} The baseline search varies two price books, two target rules, three spend packages, and two cash floors, for 24 total configurations.}
  \label{tab:rule-baseline-search-space}
\end{table}

\section{Comparison to Vending-Bench 2 Curves}
\label{app:vendingbench-comparison}

\begin{figure}[h]
  \centering
  \includegraphics[width=0.85\textwidth]{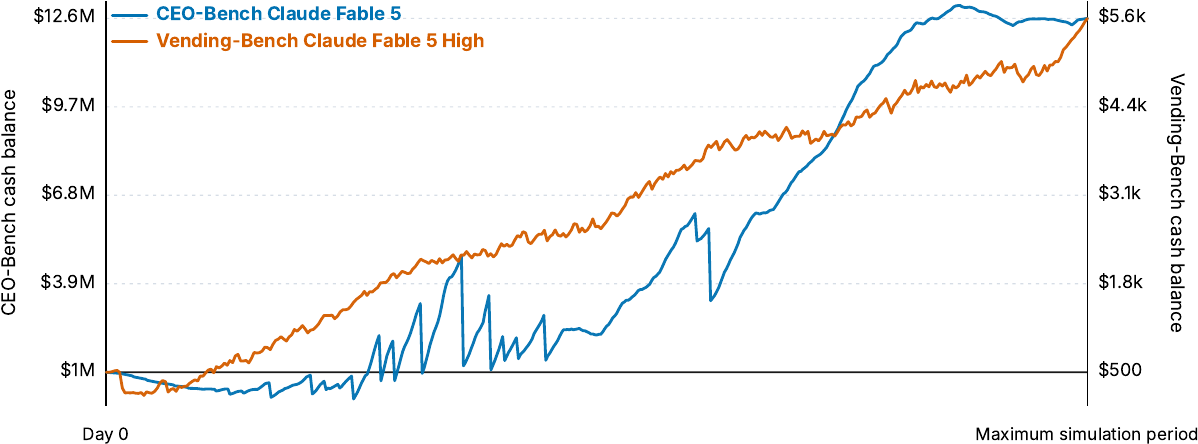}
  \caption{\textbf{Claude Fable 5 cash trajectories in Vending-Bench 2 and \textsc{CEO-Bench}.} Vending-Bench 2 reports an average Claude Fable 5 High trajectory over five 365-day runs, where cash grows from a \$500 starting balance to roughly \$5.7K. The best Claude Fable 5 \textsc{CEO-Bench} run starts at \$1M, draws down early to fund growth and product investment, and later compounds to \$12.6M, illustrating the delayed-payoff structure of our simulator.}
  \label{fig:vendingbench-comparison}
\end{figure}

Vending-Bench 2 reports the average Claude Fable 5 High balance over five 365-day runs, with the leaderboard ending near \$5.7K from a \$500 starting balance~\citep{backlund2025vendingbench2}. Its curve grows steadily after early operating volatility: once the agent finds workable suppliers and products, gains can accumulate through repeated restocking and sales. In contrast, the best Claude Fable 5 \textsc{CEO-Bench} run begins with \$1M, falls to roughly \$626K by day 30 as it funds acquisition, development, capacity, and operations, and later compounds to \$12.6M by the end of the run. The middle of the \textsc{CEO-Bench} trajectory is also more volatile, with larger repeated drawdowns and recoveries as the agent tests pricing, product investment, capacity, and acquisition policies. Exploration therefore has a direct financial cost: a strategy can be directionally correct while still producing long stretches of unstable cash balance before its benefits appear. This delayed-payoff structure makes local cash preservation an unreliable proxy for success and requires the agent to keep investing through long intervals where the benefits of its actions have not yet fully materialized.

\section{Upper Bound Final-Cash Estimate}
\label{app:upper-bound-estimation}

We use the estimated maximum final cash only to calibrate remaining headroom in \textsc{CEO-Bench}. It is not a mathematical proof of optimality. The estimate has two stages. We first compute a pre-friction accounting subtotal by summing revenue from all customer groups under maximum supportable pricing and subtracting the costs required by the simulator mechanics. We then apply a friction adjustment for issue-driven churn, enterprise negotiation friction, and acquisition delay. The reported estimate is the adjusted value, approximately \$2.2B.

\paragraph{Objective.}
Let $G$ denote the 26 customer groups and let $B=\{30,60,\ldots,480\}$ denote the billing days. For each group $g$, let $N_g$ be the maximum attainable customer count or enterprise seat count, $p_g$ be the maximum supportable price, and $r_g(t)$ be the retention probability at billing day $t$. These terms define an optimistic full-market revenue calculation: how many customers can be reached, what price they can support, and how likely they are to remain active at each renewal. The pre-friction revenue estimate is
\begin{equation}
R = \sum_{g \in G} N_g p_g \sum_{t \in B} r_g(t).
\end{equation}
Here $R$ is total subscription revenue before execution frictions. The inner sum adds the retained billing opportunities for a group over time, and the outer sum aggregates across all customer segments. This simulates the revenue side of a best-case operating plan in which acquisition succeeds and retention is governed by the modeled quality checks. The pre-friction final-cash subtotal is then
\begin{equation}
C_{\mathrm{pre}} = C_0 + R - K,
\end{equation}
where $C_{\mathrm{pre}}$ is final cash before friction adjustment, $C_0=\$1$M is the initial cash balance, and $K$ is the sum of all modeled simulator costs. This subtotal simulates accounting profit before execution slippage: revenue plus starting cash minus the cost of capacity, compute, development, operations, acquisition, and research. The reported estimate applies a friction factor $F$:
\begin{equation}
C_{\mathrm{reported}} = F C_{\mathrm{pre}}.
\end{equation}
where $C_{\mathrm{reported}}$ is the final headroom estimate after friction and $F$ is the multiplicative discount applied to the pre-friction subtotal. We use $F=0.49$ to make the estimate conservative with respect to issue-driven churn, enterprise negotiation friction, and acquisition delay. This simulates the gap between a clean accounting upper bound and a real execution path, where customers do not all arrive instantly, enterprise deals take negotiation, and support issues can erode retention.

\paragraph{Parameter choices.}
The accounting assumes all 26 groups are retained through all billing cycles before friction adjustments. The selected configuration uses T3 inference, T7 capacity, \$40K/day in global development, targeted development for every group, all ten R\&D tiers started in parallel on day 1, and operations spending of \$500/day plus \$0.001 per active subscriber. These choices were selected from a small grid over targeted-development slack, model tier, global-development spend, R\&D schedule, capacity tier, and operations spend. The most sensitive parameter is targeted development: a 1--2$\times$ multiplier did not reliably retain all groups late in the run, while a 3$\times$ multiplier maintained full-cycle retention in the tested configurations and a 4$\times$ multiplier added cost without improving retention.

\paragraph{Quality and retention check.}
For each candidate configuration, we compute the shared-quality, competitor-drift, and R\&D-bonus trajectories over the operating horizon. For group $g$, targeted-development spend is first sized to cover the late-run quality threshold,
\begin{equation}
T_g^{(0)} =
5000 \left[
\exp \left(
\frac{\Delta_g}{0.7 A_g \cdot 0.0225}
\right) - 1
\right],
\end{equation}
where $T_g^{(0)}$ is the one-pass targeted-development spend estimate for group $g$, $\Delta_g$ is the required targeted quality bonus after accounting for shared quality and R\&D effects, $A_g$ is the number of active development days, $5000$ is the spend scale from the simulator's targeted-development rule, $0.7$ is the targeted-development conversion coefficient used in this sizing approximation, and $0.0225$ is the quality-gain coefficient. The exponential is the inverse of the simulator's log diminishing-returns function: larger required quality gaps require disproportionately more spend. We then use $T_g=3T_g^{(0)}$ to account for group-level drift feedback that the one-pass sizing rule does not capture. This simulates planning with safety margin, where an operator overbudgets targeted product work because competitors and market drift can raise the bar during execution.

At each billing day, retention is computed from the delivered-quality cushion,
\begin{equation}
r_g(t) =
\Phi \left(
\frac{
m[q_0 + q_{\mathrm{shared}}(t) + b_g(t)] - d_g(t) - \ell(t) - \mu_g
}{
\sigma_g
}
\right),
\end{equation}
where $r_g(t)$ is group retention probability at billing day $t$, $\Phi$ is the standard normal cumulative distribution function, $m=0.90$ is the T3 quality multiplier, $q_0$ is baseline quality, $q_{\mathrm{shared}}(t)$ is shared quality from global development and R\&D, $b_g(t)$ is the targeted group bonus, $d_g(t)$ is fixed and competitor-induced group drift, $\ell(t)$ is the capacity-overload penalty, and $(\mu_g,\sigma_g)$ parameterize the group's quality threshold distribution. The numerator is the quality cushion above the group's threshold, and dividing by $\sigma_g$ converts that cushion into a probability under the assumed threshold distribution. Under the selected configuration, the computed retention check gives $r_g(t)\approx 1$ for all 26 groups and all 16 billing cycles. Full retention implies approximately 521M usage units per day; with T7 capacity, the resulting overload penalty is included in $\ell(t)$. This simulates a stress test for whether product quality, targeted improvements, and capacity are strong enough to keep every segment through repeated renewals.

\paragraph{Revenue and costs.}
The pre-friction calculation gives \$6.69B in subscription revenue: \$1.93B from individual groups and \$4.76B from enterprise groups. We subtract every cost category used by the simulator rather than only direct infrastructure costs. Table~\ref{tab:upper-bound-costs} shows the resulting cost accounting. This paragraph simulates the full-company accounting view of the upper bound: high revenue is meaningful only after paying for the compute, capacity, product work, support, advertising, and information acquisition required to keep that revenue.

\begin{table}[h]
  \centering
  \tablefontsize
  \begin{tabular}{l r}
    \toprule
    \textbf{Quantity} & \textbf{Amount (\$M)} \\
    \midrule
    Subscription revenue & 6{,}690.0 \\
    Initial cash & 1.0 \\
    \midrule
    Compute & 1{,}554.0 \\
    Capacity & 37.3 \\
    Global development & 19.9 \\
    Targeted development & 423.6 \\
    R\&D projects & 9.2 \\
    Advertising & 114.8 \\
    Lead acquisition & 0.9 \\
    Operations & 2.5 \\
    Market research & 1.5 \\
    Group research & 1.0 \\
    \midrule
    Total modeled costs & 2{,}164.7 \\
    Pre-friction ending cash subtotal & 4{,}526.3 \\
    Friction adjustment factor & $\times 0.49$ \\
    Estimated final cash upper bound & 2{,}200.0 \\
    \bottomrule
  \end{tabular}
  \caption{Accounting for the estimated final cash upper bound of \$2.2B after adjustments for execution frictions.}
  \label{tab:upper-bound-costs}
\end{table}

\paragraph{Interpretation.}
The final row of Table~\ref{tab:upper-bound-costs} is the estimate used in the main text and in Table~\ref{tab:results-overview}. It should be read as an approximate headroom calculation, not as a demonstrated executable strategy. The pre-friction accounting assumes full conversion, full enterprise close rates, rapid acquisition, maximum supportable prices, mean R\&D timing and quality effects, and no additional losses from unresolved customer issues, discounts, reputational effects, or cash-flow constraints beyond the cost categories above. The friction factor makes the estimate conservative by reducing this subtotal to roughly \$2.2B. This remains far above the best observed agent performance and supports the conclusion that \textsc{CEO-Bench} is far from saturated.

\end{document}